\pdfoutput=1
\documentclass[acmsmall,screen=true,nonacm=true]{acmart}

\usepackage{multirow}
\usepackage{array}
\usepackage{xurl}
\usepackage{tikz}
\usetikzlibrary{positioning,arrows.meta,fit,calc}
\usepackage{float}

\graphicspath{{figures/}}

\makeatletter
\newcommand{\ifanon}[2]{\if@ACM@anonymous#2\else#1\fi}
\makeatother

\acmVolume{}
\acmNumber{}
\acmArticle{}
\acmYear{2026}
\acmMonth{8}
\copyrightyear{2026}
\setcopyright{cc}
\setcctype{by}

\hypersetup{
  pdfpublisher={Pier-Jean Malandrino},
  pdfpubtype={preprint},
  pdfcopyright={Creative Commons Attribution 4.0 International (CC BY 4.0)},
  pdflicenseurl={https://creativecommons.org/licenses/by/4.0/}}

\begin{document}

\title{Unfolding the Leech Lattice: Fused Multi-Shell Decoding and VRAM Layouts for 2-Bit LLM Weights}
\subtitle{What a combinatorial index costs to serve, measured against deployed 4-bit and 2-bit GEMV kernels}

\author{Pier-Jean Malandrino}
\affiliation{%
  \institution{Scub}
  \city{Bordeaux}
  \country{France}}
\email{pierjean.malandrino@scub.net}

\renewcommand{\shortauthors}{Malandrino}

\begin{abstract}
Leech-lattice vector quantization holds the strongest reported 2-bit
quality under its own evaluation protocol. Its kernel decodes one shell; we found
no implementation of the multi-shell decoder the rate requires. This paper
supplies one and measures its serving cost for decode-phase GEMV at batch~1. First, a serving path
for the full 301-class codebook: an offline expansion into GPU layouts and
a fused dequantize-plus-matvec kernel reading them without warp divergence,
verified against f64. Second, the
in-VRAM rate is a design axis distinct from the on-disk rate. Four
bit-exact layouts timed in one process show binary bit planes beating
one-hot masks on size and speed at constant bandwidth (4.80 bits per
weight, $2.15\times$ FP16). Below 4.3 bits a second, irregular stream
enters; at 3.6 the decode stops being shifts and masks. Third,
deployed four-bit (AWQ) and two-bit (QTIP) GEMV kernels run in the same
process. The trellis kernel reads $2.40\times$ fewer bytes than our served
layout and runs $2.27\times$ faster at near-equal fractions of their byte
bounds: the time gap tracks the traffic gap, the price of unfolding a codebook too large for a lookup table. Fourth, the validity envelope: the trellis kernel outruns our
no-weights control, so our launch geometry sets that floor, and on a second memory hierarchy every lattice arm falls below FP16. With
the output head held identical across arms, the kernel-and-format path
gains $1.11\times$, $1.29\times$ and $1.41\times$ end to end at 4B, 8B
and 14B; with an int8 output head the served 4B reaches 87.0~tok/s in
2.60~GB. The quality cost, $\times$1.38 perplexity and 14.7 MMLU points at
4B, shrinks across the three sizes measured.
\end{abstract}



\maketitle

\section{Introduction}
\label{sec:intro}

The number of bits per weight is the only rate that changes the \emph{class}
of model a given machine can load. At 16 bits a 70B-parameter model needs
140~GB of weights; at 2 bits the same model occupies about 20~GB on disk and
fits in the memory of a workstation. That figure is the on-disk rate. The
rate served from VRAM is higher, because the kernel reads a format it can
decode inside a matvec, and that rate is the subject of \S\ref{sec:layouts}.
For deployments whose constraint is sovereignty (capable models on hardware
one owns), the 2-bit operating point decides whether the model loads at all.

The strongest reported 2-bit quality, under its own evaluation protocol,
comes from Leech-lattice vector quantization (LLVQ)~\citep{llvq2026}: weights are grouped in blocks of 24,
normalized, and snapped to the nearest point of the Leech lattice
$\Lambda_{24}$. The systems half of that paper stops short. Its CUDA kernel
decodes a single lattice shell ($m=3$, ``for simplicity''), is reported
slower than QTIP's kernel~\citep{qtip2024}, and low-level optimization is
deferred as ``largely orthogonal'' to the contribution. The decoder the
2-bit rate requires is a union of shells: 301 coordinate-permutation
classes for the ball $\Lambda_{24}(12)$, whose 47-bit index names one of
$1.1\times10^{14}$ lattice points. A branch per class would diverge across
every warp. We found no prior implementation of that decoder, the original
work included.

This paper supplies it and measures, for decode-phase GEMV at batch~1, what
a combinatorial lattice index costs to serve. Four contributions:

\begin{itemize}
\item[\textbf{C1}] \textbf{The fused multi-shell decoder}
(\S\ref{sec:decoder}): to our knowledge the first publicly documented
serving implementation
of the full multi-shell $\Lambda_{24}(12)$ codebook (301 classes): an
offline expansion into divergence-free GPU layouts, and a fused
dequantize-plus-matvec kernel that reads them with shifts, masks and a
predicated table lookup, the same instruction sequence on every lane
whatever the class. The combinatorial index itself is never decoded inside
the matvec. Every output row is verified against an f64 reference before
any timing; on an L40S the kernel reaches 65\% of its byte bound, measured
against the FP16 control's bandwidth (\S\ref{sec:layouts}).

\item[\textbf{C2}] \textbf{The in-VRAM rate as a design axis}
(\S\ref{sec:layouts}), distinct from the on-disk rate (2.07 effective,
\S\ref{sec:evaluation}). Four bit-exact layouts timed in one process span
5.51, 4.80, 4.34 and 3.59 bits per weight. Binary bit planes beat one-hot
masks on size and speed at constant bandwidth: Planes14, the served layout,
reads 4.80 bits per weight at $2.15\times$ an FP16 matvec. The curve turns
between 4.8 and 4.3 bits per weight, where a second, irregular stream (the
exception side channel) enters, and collapses at 3.6, where the decode stops
being shifts and masks.

\item[\textbf{C3}] \textbf{The cost of a combinatorial index, located}
(\S\ref{sec:qtip}): a same-process comparison with deployed 4-bit
(AWQ w4g128) and 2-bit (QTIP) GEMV kernels. The trellis kernel reads
$2.40\times$ fewer bytes than Planes14 (0.91 against 2.18~GB) and runs
$2.27\times$ faster. Kernel efficiency is close (61\% against 65\% of
bound); in this comparison the time gap tracks the byte ratio. As shipped, launch geometry is not
equalized between the two kernels; \S\ref{sec:attribution} recovers 11.7\%
of Planes14's time by fusion, so the byte-driven residual is somewhat under
$2.27\times$. A codebook of $1.1\times10^{14}$ points
cannot live in a lookup table, where E8P- and IQ2-style codebooks hold at
most $2^{16}$ entries, so it is unfolded at load time into a 4.80-bit
stream.

\item[\textbf{C4}] \textbf{The validity envelope, measured}
(\S\ref{sec:attribution}, \S\ref{sec:a100}, \S\ref{sec:integration}). A
no-weights control runs the same launches over the same shapes and reads no
weight bytes; the trellis kernel outruns it, so our launch geometry and not
the card sets that time. An attribution of the gap to the DRAM floor places
39\% in launch geometry, and fusing q/k/v and gate/up (144 launches instead
of 252) recovers 11.7\% of the Planes14 kernel time and $1.061\times$ end to
end on the served path. On a second memory hierarchy (A100) every lattice arm
falls below FP16. End-to-end serving is measured on three model sizes.
\end{itemize}

With the output head held identical in both arms, the kernel-and-format
path gains $1.11\times$, $1.29\times$ and $1.41\times$ end to end at 4B,
8B and 14B; with an int8 output head the served 4B configuration reaches
87.0~tok/s in 2.60~GB of VRAM (\S\ref{sec:integration}). At 4B, two bits cost $\times1.384$ perplexity
and 14.7 MMLU points where a 4-bit baseline keeps full quality; the deficit
shrinks at 8B and 14B, and three points support no scaling law
(\S\ref{sec:evaluation}, Appendix~\ref{app:scale}). Every figure
regenerates from CSV files committed with the source, and the tables are
checked against those files on every build.

\section{The Fused Multi-Shell Decoder}
\label{sec:decoder}

\subsection{What a $\Lambda_{24}$ index is}

LLVQ quantizes weights in blocks of 24. Each block is normalized and its
direction is snapped to the nearest point of the Leech lattice
$\Lambda_{24}$ inside the ball $\Lambda_{24}(12)$ (squared norm
$\le 2\cdot 12$ in the standard scaling). The ball holds
$N = 111\,043\,117\,458\,000$ points, so a direction index costs
$\lceil\log_2 N\rceil = 47$ bits; one gain bit selects one of two
per-tensor magnitude centroids. A block costs 48 bits, a code rate of
2.000~b/weight.

In $\sqrt{8}\,\Lambda_{24}\subset\mathbb{Z}^{24}$ a lattice point satisfies
three conditions tied to the binary Golay code $\mathcal{G}_{24}$
\citep{conway1999sphere}: all coordinates even or all odd, a codeword
supported by the coordinates in a fixed residue class mod~4, and a sum
condition mod~8. Every point of the ball falls in one of 301
\emph{classes}, the orbits under coordinate permutation and admissible sign
changes, each fixed by a codeword weight, a magnitude multiset and a coset.
Format~v1 enumerates classes, then arrangements and signs within a class, in
a fixed order. The index is a bijection, and a test pins a fingerprint of
the whole map down to its mixed-radix composition orders. \emph{Full
multi-shell} here means all 301 classes of the $\Lambda_{24}(12)$ ball, not
an enumeration of its $1.1\times10^{14}$ points as a lookup table.

\subsection{Unfolding at load time}

The encoder (nearest-neighbor search over $1.1\times10^{14}$ points) runs
offline, once per model, and may cost hours; the decoder runs inside every
matvec. At the 2-bit rate the codebook spans eleven shells and 301 classes,
and a per-class branch would diverge across every warp. The original
paper's kernel decodes a single shell ($m=3$).

\begin{figure}[t]
\centering
\includegraphics[width=\linewidth]{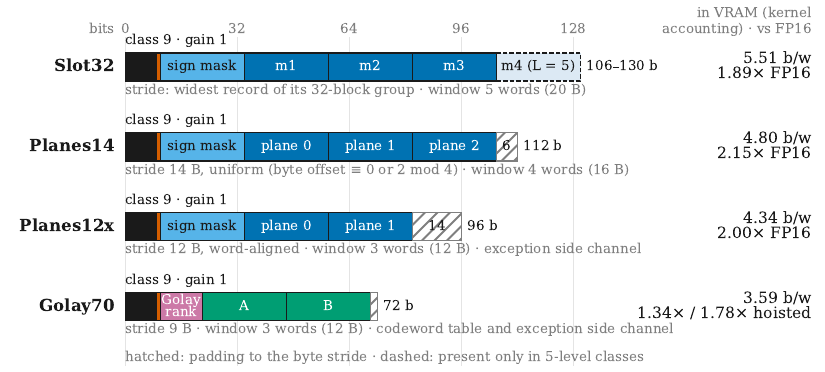}
\Description{Four horizontal bit-scale byte maps, one per layout, showing
the fields of each decoded record with their widths in bits, the stride and
read window under each, and the in-VRAM rate and speedup over FP16 at
right.}
\caption{The four decoded-field records to bit scale, with field widths,
stride, and read window in 32-bit words. \textsc{Slot32}: 106 bits at $L=4$
to 130 at $L=5$, five-word window, a per-group base table as side channel.
\textsc{Planes14}: 112 bits at a uniform 14-byte stride, four-word window,
no side channel. \textsc{Planes12x}: 96 bits in three aligned words, plus a
per-matrix exception table holding the exact \textsc{Planes14} record of
every five-level block (3.38\% of blocks), applied in the same launch.
\textsc{Golay70}: 72 bits, a 16~KiB codeword table ($4\,096 \times 24$
bits), and an exception table for the blocks its in-class 2-bit code cannot
hold (7.44\%). Right: kernel b/weight and speed against FP16; the
\textsc{Golay70} speed is the hoisted decoder's (Table~\ref{tab:layouts}).
Data: echelle-formats.csv.}
\label{fig:records}
\end{figure}

Our design choice is that the 47-bit index is never decoded in the matvec.
An offline \emph{transcoder} expands each block once, at load time, into a
record of decoded fields: class identifier, gain bit, sign mask, and the
magnitude level of each coordinate, as one-hot masks (\textsc{Slot32}) or
as binary bit planes (\textsc{Planes14}). Figure~\ref{fig:records} draws
the four records of \S\ref{sec:layouts} to bit scale. The kernel reads a
record with shifts and masks, the same instruction sequence for every lane
whatever its class. That buys a divergence-free decode, and it costs the
in-VRAM rate of \S\ref{sec:layouts}: the unfolded \textsc{Planes14} record
is 112 bits wide where the index it replaces is 48.

\subsection{The kernel}

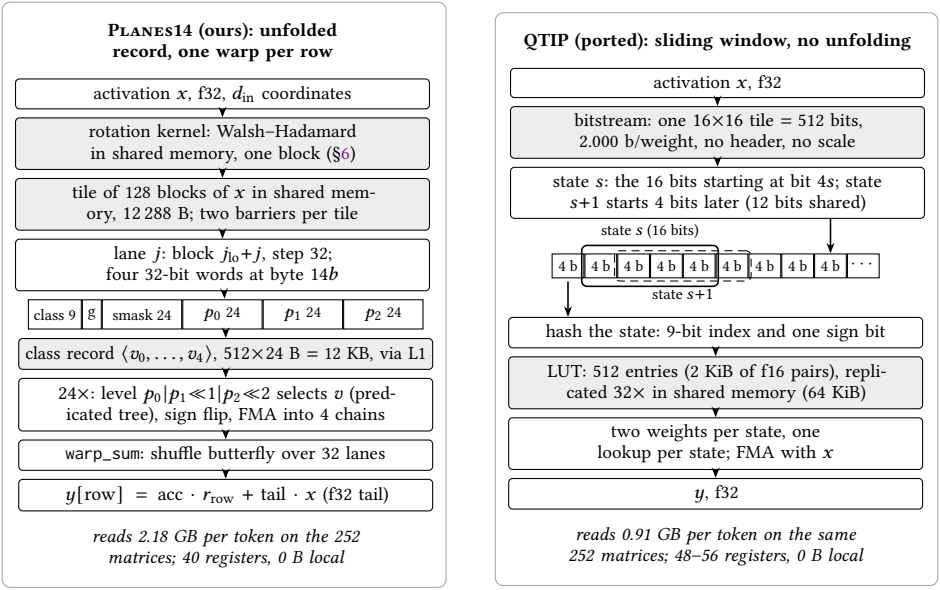
\begin{figure}[t]
\centering
\begin{tikzpicture}[
  font=\scriptsize,
  node distance=3pt and 8pt,
  box/.style={draw, rounded corners=1.5pt, inner sep=2.5pt, align=center,
              text width=5.3cm, minimum height=11pt},
  mem/.style={box, fill=black!7},
  field/.style={draw, inner sep=0pt, minimum height=11pt, font=\tiny,
                anchor=north west},
  cell/.style={draw, inner sep=0pt, minimum width=0.42cm, minimum height=9pt,
               font=\tiny, anchor=north west},
  flow/.style={-{Stealth[length=4pt]}, semithick},
  title/.style={font=\scriptsize\bfseries, align=center, text width=5.3cm},
  note/.style={font=\scriptsize\itshape, align=center, text width=5.3cm},
  panel/.style={draw, black!35, rounded corners=3pt, inner sep=4pt},
]
\node[title] (lt) {\textsc{Planes14} (ours): unfolded record, one warp per row};
\node[box, below=of lt] (x) {activation $x$, f32, $d_{\mathrm{in}}$ coordinates};
\node[mem, below=of x] (rot) {rotation kernel: Walsh--Hadamard in shared
  memory, one block (\S\ref{sec:limitations})};
\node[mem, below=of rot] (tile) {tile of 128 blocks of $x$ in shared memory,
  $12\,288$~B; two barriers per tile};
\node[box, below=of tile] (lane) {lane $j$: block $j_{\mathrm{lo}}{+}j$, step
  32; four 32-bit words at byte $14b$};
\node[field, minimum width=0.70cm] (r1) at ([xshift=0.17cm, yshift=-3pt]lane.south west) {class 9};
\node[field, minimum width=0.25cm, right=0pt of r1.north east, anchor=north west] (r2) {g};
\node[field, minimum width=1.05cm, right=0pt of r2.north east, anchor=north west] (r3) {smask 24};
\node[field, minimum width=1.05cm, right=0pt of r3.north east, anchor=north west] (r4) {$p_0$ 24};
\node[field, minimum width=1.05cm, right=0pt of r4.north east, anchor=north west] (r5) {$p_1$ 24};
\node[field, minimum width=1.05cm, right=0pt of r5.north east, anchor=north west] (r6) {$p_2$ 24};
\node[fit=(r1)(r6), inner sep=0pt] (rec) {};
\node[mem, below=of rec] (tab) {class record $\langle v_0,\dots,v_4\rangle$,
  $512\times24$~B $=$ 12~KB, via L1};
\node[box, below=of tab] (fma) {$24\times$: level $p_0|p_1{\ll}1|p_2{\ll}2$
  selects $v$ (predicated tree), sign flip, FMA into 4 chains};
\node[box, below=of fma] (ws) {\texttt{warp\_sum}: shuffle butterfly over 32 lanes};
\node[box, below=of ws] (ep) {$y[\mathrm{row}] = \mathrm{acc}\cdot r_{\mathrm{row}}
  + \mathrm{tail}\cdot x$ (f32 tail)};
\node[note, below=of ep] (lb) {reads 2.18~GB per token on the 252 matrices;
  40 registers, 0~B local};
\draw[flow] (x) -- (rot);
\draw[flow] (rot) -- (tile);
\draw[flow] (tile) -- (lane);
\draw[flow] (lane) -- (rec);
\draw[flow] (rec) -- (tab);
\draw[flow] (tab) -- (fma);
\draw[flow] (fma) -- (ws);
\draw[flow] (ws) -- (ep);
\node[panel, fit=(lt)(x)(ep)(lb)] {};
\node[title, right=1.0cm of lt] (rt) {QTIP (ported): sliding window, no unfolding};
\node[box, below=of rt] (qx) {activation $x$, f32};
\node[mem, below=of qx] (qbits) {bitstream: one $16{\times}16$ tile $=$ 512
  bits, 2.000~b/weight, no header, no scale};
\node[box, below=of qbits] (qwin) {state $s$: the 16 bits starting at bit
  $4s$; state $s{+}1$ starts 4 bits later (12 bits shared)};
\node[cell] (c1) at ([xshift=0.55cm, yshift=-13pt]qwin.south west) {4 b};
\node[cell, right=0pt of c1.north east, anchor=north west] (c2) {4 b};
\node[cell, right=0pt of c2.north east, anchor=north west] (c3) {4 b};
\node[cell, right=0pt of c3.north east, anchor=north west] (c4) {4 b};
\node[cell, right=0pt of c4.north east, anchor=north west] (c5) {4 b};
\node[cell, right=0pt of c5.north east, anchor=north west] (c6) {4 b};
\node[cell, right=0pt of c6.north east, anchor=north west] (c7) {4 b};
\node[cell, right=0pt of c7.north east, anchor=north west] (c8) {4 b};
\node[cell, right=0pt of c8.north east, anchor=north west] (c9) {4 b};
\node[cell, right=0pt of c9.north east, anchor=north west] (c10) {$\cdots$};
\node[draw, semithick, rounded corners=2pt, inner xsep=1pt, inner ysep=3pt,
      fit=(c2)(c5), label={[font=\tiny, yshift=-1pt]above:state $s$ (16 bits)}] (w1) {};
\node[draw, densely dashed, rounded corners=2pt, inner xsep=1pt, inner ysep=1pt,
      fit=(c3)(c6), label={[font=\tiny, yshift=1pt]below:state $s{+}1$}] (w2) {};
\node[fit=(c1)(c10)(w1)(w2), inner sep=0pt] (strip) {};
\node[box, below=11pt of strip] (qhash) {hash the state: 9-bit index and one sign bit};
\node[mem, below=of qhash] (qlut) {LUT: 512 entries (2~KiB of f16 pairs),
  replicated $32\times$ in shared memory (64~KiB)};
\node[box, below=of qlut] (qfma) {two weights per state, one lookup per
  state; FMA with $x$};
\node[box, below=of qfma] (qy) {$y$, f32};
\node[note, below=of qy] (rb) {reads 0.91~GB per token on the same 252
  matrices; 48--56 registers, 0~B local};
\draw[flow] (qx) -- (qbits);
\draw[flow] (qbits) -- (qwin);
\draw[flow] (qwin.south -| c9.north) -- (c9.north);
\draw[flow] (c1.south |- w2.south) -- (c1.south |- qhash.north);
\draw[flow] (qhash) -- (qlut);
\draw[flow] (qlut) -- (qfma);
\draw[flow] (qfma) -- (qy);
\node[panel, fit=(rt)(qx)(qy)(rb)] {};
\end{tikzpicture}
\Description{Two flowcharts side by side: the Planes14 kernel's path from
the activation through the rotation, the shared-memory tile, the four-word
record window, the class table, the fused multiply-adds and the warp sum to
the output row; and the ported QTIP kernel's path through a 16-bit sliding
window, a hash, a 512-entry lookup table and the multiply-adds.}
\caption{Two decodes of a 2-bit weight stream on the same matvec shapes.
Left: one warp of the \textsc{Planes14} kernel, activation to output row;
shaded boxes are memory-resident objects. Right: the QTIP kernel as ported,
a 16-bit state sliding over the stored bits and a lookup in place of an
unfolded record. Byte counts are one token over the 252 projections of
Qwen3-4B; register counts are the compiler's. Data: echelle-formats.csv.}
\label{fig:dataflow}
\end{figure}

Figure~\ref{fig:dataflow} (left) follows one warp through the
\textsc{Planes14} kernel. One warp computes one output row; a 256-thread CTA
holds 8 rows and the grid holds $d_{\mathrm{out}}/8$ CTAs. A separate
kernel rotates the activation first, a Walsh--Hadamard transform in the
shared memory of a single block (the bound of \S\ref{sec:limitations}). The
matvec stages $x$ in shared memory by tiles of 128 blocks ($12\,288$
bytes), two barriers per tile; within a tile, lane $j$ owns block
$j_{\mathrm{lo}}+j$ and steps by 32.

A lane reads its record through a four-word window: the record starts at
byte $14b$, which is $\equiv 0$ or $2 \pmod 4$, so the in-word shift is 0 or
16 bits and $16+106$ payload bits fit in 128 (\textsc{Slot32} needs five
words). The level of coordinate $i$ is
$p_0[i]\,|\,p_1[i]{\ll}1\,|\,p_2[i]{\ll}2$. It selects one of the class's
at most five values through a predicated tree rather than a dynamically
indexed array, so the kernel uses no local memory. A sign flip and a fused
multiply-add on $x[i]$ follow, into one of four independent accumulator
chains. After the last tile a shuffle butterfly sums the warp, and lane 0
writes $y = \mathrm{acc}\cdot r + \mathrm{tail}\cdot x$, with $r$ the row
scale and the tail the columns past the last full block, kept in f32
(KeepExact).

The class table holds 512 records of 24 bytes (five f32 levels and a count),
12~KB, sized above the 301 classes so a 9-bit identifier cannot index out
of bounds, and read through L1 as a \texttt{const}-qualified global pointer.
Constant memory would serialize the 32 distinct addresses a warp issues;
the gather costs 0.041~ms per token (\S\ref{sec:attribution}).
Every point of a shell has the same norm, so the normalization to a unit
direction is a per-class constant folded into the stored levels, and the
multi-shell rescaling of the original paper's Appendix~G is one multiply per
block. The compiler reports 40 registers and no local memory for the kernel; the
controls and competitors span 31--56 (Table~\ref{tab:fairness}).

The right panel of Figure~\ref{fig:dataflow} draws the 2-bit competitor's
decode. A 16-bit state slides over the bitstream by 4 bits per step, so
consecutive states overlap by 12 bits; one hash and one lookup in a
512-entry table yield two weights. Nothing is unfolded, and on the same 252
matrices it reads 0.91~GB where \textsc{Planes14} reads 2.18
(\S\ref{sec:qtip}).

\subsection{Verification}

Lattice decoders fail silently: a wrong sign convention or a swapped
enumeration order yields plausible weights and a ruined model. Three
defenses:

\begin{itemize}
\item \textbf{Bit-exact transcoding.} The transcoder from format~v1 to the
four byte formats of \S\ref{sec:layouts} is verified bit-exact on all
$150\,681\,600$ blocks of the artifact, its tests hardened by mutation
testing (${\sim}25$ mutants killed).

\item \textbf{Every row against f64 before any timing.} Each benchmark
checks every output row of every arm ($1\,105\,920$ rows) against an
independent f64 reference at $10^{-5}\cdot\sum|w\cdot x|$; our arms measure
$2.2$ to $3.4\times10^{-8}$. The two arms storing a binary16 output (the
4-bit competitor and cuBLAS) are held to $10^{-3}$ and measure
$5.8\times10^{-5}$ and $5.7\times10^{-5}$.

\item \textbf{Invariant locks upstream.} The lattice code is pinned by exact
combinatorial invariants: the kissing number $196\,560$, the theta-series
coefficients, and the class-cardinality sum
$N(13)=280\,974\,212\,784\,720$, which no incorrect enumeration constraint
reproduces.
\end{itemize}

\section{VRAM Layouts: the In-VRAM Rate as a Design Axis}
\label{sec:layouts}

The rate that decides which models fit on a GPU is the rate of the format
the kernel reads. It is a different quantity from the on-disk rate
(2.07~b/weight effective, \S\ref{sec:evaluation}). This section maps it on
one card: four byte formats of ours read by five decoders, a deployed 4-bit
and a deployed 2-bit kernel, FP16 and cuBLAS controls, and a control kernel
that reads no weights. The ten arms share one process and one
byte-accounting convention.

\subsection{Protocol}
\label{sec:protocol}

All arms run in one process on one NVIDIA L40S: one token through the 252
projections of Qwen3-4B (decode-phase GEMV, batch~1). The ten arms are
interleaved in a fixed dispatch order over 7 rounds, the first 2 discarded.
Every speedup is the median of per-round ratios against the FP16 control,
with its min--max range. Every output row of every arm is verified against
f64 before any timing (\S\ref{sec:decoder}). Byte accounting is the
kernel's: what each arm streams per matrix (payload, exception side
channels, f32 tail columns, per-row scales). The 4-bit competitor bills no
tail, which understates our arms. Effects smaller than a range are not
separated, and second decimals read as dispersion. Acceptance criteria for
each layout were fixed and timestamped before measuring; their list, the
run's two phases, and a per-arm table of what is and is not held equal
(payload provenance, grid, output type, billing, verification threshold)
are in Appendix~\ref{app:protocol} (Table~\ref{tab:fairness}).

\subsection{The layout scale}
\label{sec:scale}

\begin{table}[t]
\centering
\small
\setlength{\tabcolsep}{3pt}
\begin{tabular}{lrrrrrrl}
\toprule
Kernel & b/weight & GB & med.\ ms & GB/s & of bound & vs FP16 & status \\
\midrule
FP16 (control) & 16.000 & 7.27 & 10.994 & 661 & --- & 1.00$\times$ & \\
FP16 via cuBLAS & 16.000 & 7.27 & 10.830 & 672 & 102\% & 1.02$\times$ [1.02--1.02] & checks the control \\
\midrule
\textsc{Slot32} & 5.510 & 2.50 & 5.820 & 431 & 65\% & 1.89$\times$ [1.89--1.89] & baseline layout \\
\textsc{Planes14} & 4.804 & 2.18 & 5.103 & 428 & 65\% & \textbf{2.15$\times$} [2.15--2.16] & \textbf{served} \\
\textsc{Planes12x} & 4.342 & 1.97 & 5.492 & 359 & 54\% & 2.00$\times$ [2.00--2.00] & best exact rate \\
\textsc{Golay70} & 3.589 & 1.63 & 8.187 & 199 & 30\% & 1.34$\times$ [1.34--1.34] & below criterion \\
\textsc{Golay70}, hoisted & 3.589 & 1.63 & 6.182 & 264 & 40\% & 1.78$\times$ [1.77--1.78] & below criterion \\
\midrule
AWQ w4g128 & 4.179 & 1.90 & 3.252 & 584 & \textbf{88\%} & 3.38$\times$ [3.37--3.38] & 4-bit competitor \\
QTIP 2-bit & 2.000 & 0.91 & \textbf{2.246} & 405 & 61\% & 4.89$\times$ [4.89--4.90] & \textbf{2-bit competitor} \\
\midrule
No-weights control & 0.159 & 0.07 & 2.306 & 31 & 5\% & 4.77$\times$ [4.76--4.77] & our launch geometry \\
\bottomrule
\end{tabular}
\caption{The layout scale: one token through the 252 projections of
Qwen3-4B, one L40S, ten arms in one process, median of per-round ratios
with min--max ranges. \emph{Of bound} is achieved bandwidth as a fraction
of the control's 661~GB/s; a ratio against FP16 is mechanically higher for
the arm reading least. The control is our own FP16 matvec; cuBLAS on the
same weights runs $1.02\times$ it. The no-weights control runs the same
252 launches and reads no weight bytes; its 0.07~GB are the f32 tail columns
and row scales the accounting bills on every arm of ours. Kernel accounting:
whole-model bits per parameter (\S\ref{sec:evaluation}) are a different
denominator. Data:
\texttt{echelle-formats.csv}.}
\label{tab:layouts}
\end{table}

\begin{figure}[t]
\centering
\includegraphics[width=0.74\linewidth]{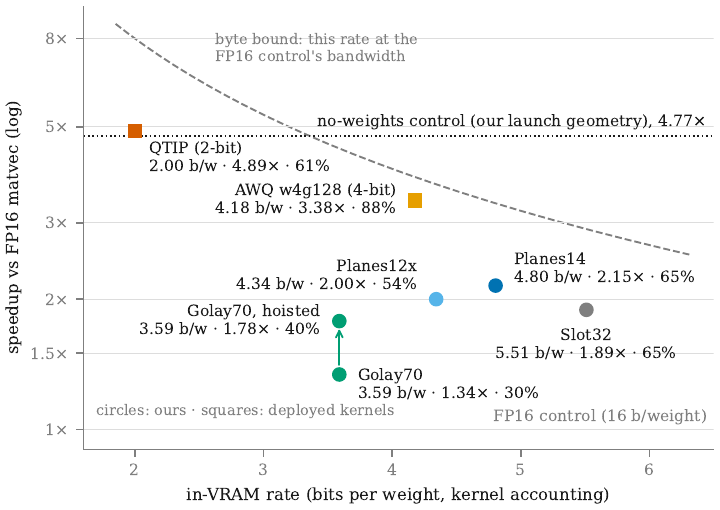}
\Description{Scatter plot of speedup over FP16 on a log axis against
in-VRAM bits per weight for the ten arms, with a dashed byte-bound
hyperbola, a horizontal rule for the no-weights control and an arrow from
the Golay70 decoder to its hoisted version.}
\caption{Speedup over FP16 against in-VRAM rate; min--max ranges over kept
rounds are smaller than the markers (Table~\ref{tab:layouts}). The dashed
hyperbola is the byte bound: the speedup an
arm at that rate would reach at the FP16 control's bandwidth on the same
shapes. An arm's fraction of its byte bound is its height as a proportion
of that curve, hence the log axis. The horizontal rule is the no-weights
control, our launch geometry over the same 252 launches. The arrow is the
hoisted \textsc{Golay70} decoder at unchanged stored bytes. Data:
\texttt{echelle-formats.csv}.}
\label{fig:layouts}
\end{figure}

Table~\ref{tab:layouts} carries the ten arms, and Figure~\ref{fig:layouts}
plots them against the byte bound, the speedup $16/b$ that an arm at rate
$b$ would reach at the control's bandwidth. Its \emph{of bound} column,
achieved bandwidth as a fraction of the control's, is the quantity that
compares across rates.

\textbf{\textsc{Slot32}} stores decoded fields at fixed offsets in a
byte-aligned record of $9 + 1 + 24L$ bits: class identifier, gain bit,
24-bit sign mask and one 24-bit one-hot mask per non-zero magnitude level.
The kernel reads a fixed five-word
(20~B) window per block, an alignment the format guarantees. It is the
simplest divergence-free format, at $1.89\times$ over FP16. At
5.510~b/weight it occupies more memory than a 4-bit format.

\textbf{\textsc{Planes14}} replaces the one-hot level masks by three binary
bit planes: a block carries three to five magnitude levels, so three bits
name each coordinate's level where one-hot spent up to four 24-bit masks.
The record is 112 bits at a uniform 14-byte stride, read through a
four-word window, with no level cap and no base pointers. It shrinks to
4.804~b/weight and runs $1.14\times$ faster than \textsc{Slot32} at
bit-identical decoded content [1.13--1.15 across jobs], $2.15\times$
over FP16. Bandwidth is unchanged, 431 to 428~GB/s: time falls as bytes
do, and the record is not what binds (\S\ref{sec:attribution}).

\textbf{\textsc{Planes12x}} adds a sparse overlay. Blocks with at most four
levels take a 12-byte two-plane record; the 3.38\% of blocks with five
levels keep their exact 14-byte \textsc{Planes14} record in a per-matrix
exception table, and extra CTAs of the same launch add the correction
$(\mathrm{exact} - \mathrm{approx}) \cdot x$. Reconstruction is bit-exact.
The alternative is not free: hard-capping the level count at four (a plain
12-byte format at about 4.1~b/weight) costs $+4.75\%$ perplexity, enough
to fall behind QTIP. The overlay reaches 4.342~b/weight at $2.00\times$
with no quality change. Its fraction of the byte bound falls from 65\% to
54\%, and the record is still shifts and masks: the cost is the second,
irregular stream, the exception table and the correction CTAs that apply
it.

\textbf{\textsc{Golay70}} removes the sign mask. Signs on the codeword
support are determined up to the Golay coset, so a 12-bit codeword rank
replaces the 24 sign bits: a 72-bit record at a 9-byte stride,
3.589~b/weight. Reconstruction is exact on all 150{,}681{,}600 blocks of
the artifact, with an exception table for the 7.44\% of blocks the
in-class two-bit code cannot hold. Recovering the signs determines the
coset per slot, and the kernel stops being bandwidth-bound: 199~GB/s, 30\% of
its byte bound, $1.34\times$, below the $1.6\times$ criterion of
Appendix~\ref{app:protocol}. A second decoder hoists the coset logic into a
per-block prologue of about a dozen integer operations, leaving three mask
tests and a negation per slot, at unchanged stored bytes. It reaches
264~GB/s, 40\% of bound and $1.78\times$, below the $2.0\times$ criterion
of Appendix~\ref{app:protocol} (the speed of \textsc{Planes12x}). Both
decoders remain points on the curve rather than served layouts.

The scale is nonlinear. Bits saved are time saved while the kernel is
bandwidth-bound on this card (\S\ref{sec:a100}) and the decode that saves
them stays shifts and masks over one stream.
For lattice codes here the curve turns between 4.8 and 4.3~b/weight, where
a second, irregular stream (the exception side channel) enters, and
collapses at 3.6, where the decode stops being shifts and masks.

These four are the survivors of a wider sweep, and the floor of the
shifts-and-masks regime is measured rather than assumed. A variable-width
layout reaches 2.40~b/weight and reads 1.09~GB against Planes14's 2.18 on
the same matrices, half the traffic, its output bit-exact against the f64
reference on all 1{,}105{,}920 rows. In a separate run at a pinned commit
its decode runs at 25~GB/s, 0.25$\times$ FP16, a factor 6.4 under the
1.6$\times$ floor: two binomial walks, a Golay word, a parity repair and
three sign rules are the arithmetic the served layouts avoid. Two further
routes below Planes14 stay closed for the decode rather than the bytes.
Transposing the planes to a byte-sliced record removes the byte rounding
but grows the record once warp-aligned. Decoding the on-disk index in the
kernel (3.04~b/weight) reintroduces the data-dependent serial decode the
load-time transcoder removes. The four served layouts occupy the rates
where a byte saved is still time saved.

\subsection{Where deployed 4-bit and 2-bit kernels sit}
\label{sec:qtip}

\textbf{The 4-bit arm} is the AWQ w4g128 GEMV kernel~\citep{awq2024},
ported unchanged into the harness with its own grid and its own binary16
output, and timed in the same rounds as ours. Its payload is synthesized
from our weights requantized to w4g128: sound for timing, since the kernel
has no data-dependent branch, and silent on quality. It reads
4.179~b/weight, less than three of our four formats, so its $3.38\times$
against FP16 is not the comparable quantity. The fraction of the byte bound
is: \textbf{88\% for the 4-bit kernel against 65\% for our best layout}.
w4g128 quantizes every column and bills no tail; billing \textsc{Planes14}
without tail and row scales gives 63\% rather than 65\%, widening the gap.
We keep 65\%, what the served format streams.

\textbf{The 2-bit arm} is the published QTIP inference
kernel~\citep{qtip2024}, fetched at a pinned commit (GPL~v3, measured and
not redistributed), run in its own launch geometry as shipped, nothing
tuned in either direction. Its payload is pseudo-random, sound for a
fixed-rate code read by a kernel without data-dependent branches, and it
claims no quality. All $1\,105\,920$ output rows pass the f64 check at our
tolerance of $10^{-5}$, worst error $5.4\times10^{-8}$. In the same phase,
process and shapes, the QTIP kernel runs $\mathbf{2.27\times}$ [2.27--2.28]
faster than \textsc{Planes14}. This is the one cross-arm ratio in the
paper: the bench prints ratios against FP16, so it is formed from the two
medians, with its range widened outward by both arms' min--max.

\textbf{The mechanism is bytes.} Both formats store 2.000 bits of code per
weight on disk, and they are comparable in quality in their respective harnesses at matched code rate
(Table~\ref{tab:lit}, \S\ref{sec:evaluation}). On the same matrices the
trellis kernel reads 0.91~GB where \textsc{Planes14} reads 2.18: a
$2.40\times$ byte ratio for a $2.27\times$ time ratio, the two kernels
converting 61\% and 65\% of their byte bounds. At near-equal efficiency
the time gap tracks the traffic gap, and what sets the traffic is the
load-time unfolding of \S\ref{sec:decoder}. A codebook
of $1.1\times10^{14}$ points cannot sit in a lookup table, where a 16-bit
trellis state can (a 2~KiB LUT; table codebooks such as QuIP\#'s E8P hold
$2^{16}$ entries). The lattice index is therefore unfolded into a
4.80~b/weight stream of bit planes, and the kernel pays for those bytes at
memory speed.

\textbf{The no-weights control.} The last row of Table~\ref{tab:layouts}
runs the same 252 launches over the same shapes and reads no weight bytes:
2.306~ms. The QTIP kernel finishes the same 252 projections, reading
0.91~GB, in 2.246~ms, 2.6\% less against a 0.36\% resolution. That control
therefore bounds our launch geometry, one warp per output row across 252
launches; a differently shaped kernel passes under it.

The fraction of the byte bound compares an arm to $16/b$ times the control,
and it is readable only while $16/b$ stays below the no-weights control,
that is while $b > 16/4.77 = 3.35$~b/weight. At 2.000~b/weight the QTIP
byte bound is $8.00\times$ against a control at $4.77\times$, so the metric
saturates there, and at 2 bits the time is the comparable quantity.

\subsection{Attribution to the DRAM floor}
\label{sec:attribution}

\begin{figure}[t]
\centering
\includegraphics[width=\linewidth]{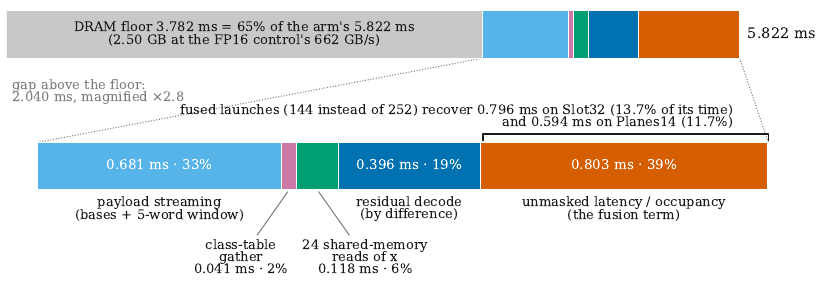}
\Description{Two stacked horizontal bars: the Slot32 time of 5.822 ms as a
DRAM floor of 3.782 ms plus five attributed terms, and the 2.04 ms gap
magnified with each term labelled and the fusion recovery bracketed over the
latency term.}
\caption{Attribution of the 2.04~ms between \textsc{Slot32} (5.82~ms) and
its DRAM floor (3.78~ms at the control's 662~GB/s): a decomposition by
subtraction of control kernels rather than a hardware profile. The
residual-decode term is by difference. CUDA events bound the host share at
0.1--0.2\% of the wall on every arm, so the latency term is gaps on the
stream rather than host submission. The marked segment is the term recovered
by fusing $q/k/v$ and gate/up. Data: \texttt{attribution-slot32.csv}.}
\label{fig:attribution}
\end{figure}

The instrumented job carries its own FP16 control, 7.27~GB at 662~GB/s (a
rounding step from Table~\ref{tab:layouts}'s 661; floor and time come from
one process). At that bandwidth the 2.50~GB that \textsc{Slot32} reads put
its DRAM floor at 3.78~ms, while the instrumented arm takes 5.82~ms. The
\textbf{2.04~ms} gap belongs to the kernel. Three control kernels, each
adding one stage, and one occupancy experiment attribute it
(Figure~\ref{fig:attribution}). Unmasked latency and occupancy take
0.803~ms (39\%) and payload streaming through the five-word window
0.681~ms (33\%); residual decode, the 24 shared-memory reads of $x$ and
the class-table gather take 0.396, 0.118 and 0.041~ms. The occupancy
experiment shows the mechanism shape by shape. The L40S holds about 852
resident CTAs (142 SMs $\times$ 6). The $k$ and $v$ projections
($d_{\mathrm{out}} = 1024$) launch 128 of them, a 15\% fill, and reach
157~GB/s against 469 for the best-filled shape ($d_{\mathrm{out}} = 9728$),
$1.06\times$ FP16 on those two shapes.

The largest term is recovered by launch geometry alone. Fusing $q/k/v$ and
$\textit{gate}/\textit{up}$ into row-concatenated launches takes the model from
252 kernels to 144, and returns 0.796~ms (13.7\%) on \textsc{Slot32} and
0.594~ms [0.586--0.596] (11.7\%) on \textsc{Planes14} in the process of
Table~\ref{tab:layouts}. Output is bit-identical on the $921\,600$ rows the
fusion touches ($o$ and down are unchanged) and bytes read are identical. No
ratio against FP16 is formed from it: the FP16 arm has no fused counterpart.

That geometry is on the served path, where it is worth $1.061\times$
[1.050--1.069] end to end on the 4B. Both arms run in one process from one
NVRTC translation unit, differing only in the launch count; the rotation hoist
is held on in both, a fused group being one site. Correctness is the gate
rather than a tolerance: the two arms emit the same 128 greedy tokens and
diverge from the dense engine at the same token 89. The table naming each row's
gain centroids costs four bytes a fused row, 0.008~b/weight.
Table~\ref{tab:e2e} holds one configuration across the three sizes and reports
the unfused path; only the 4B was re-timed fused.

The $2.15\times$ of Table~\ref{tab:layouts} is therefore not a bound
set by the format. Its largest term is launch geometry, orthogonal to the record
format, and 11.7\% of the \textsc{Planes14} time is already recovered.
(\S\ref{sec:qtip} reads the same from the two competitors.)

\subsection{A second memory hierarchy}
\label{sec:a100}

\begin{figure}[t]
\centering
\includegraphics[width=\linewidth]{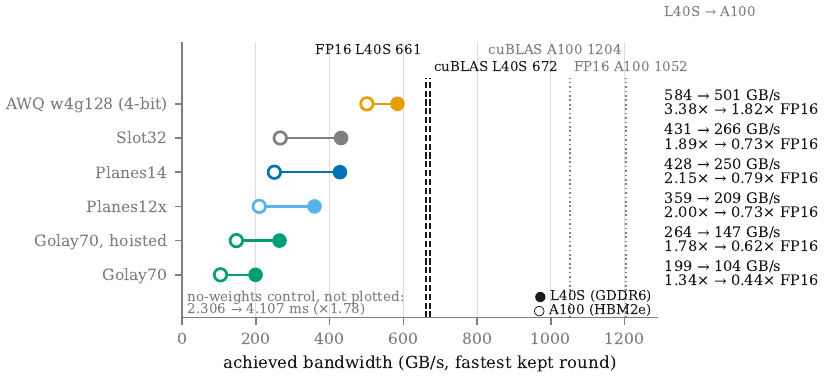}
\Description{Dumbbell chart of achieved bandwidth in GB/s for six arms,
a filled marker for the L40S and a hollow one for the A100 joined by a
line, with vertical rules for the FP16 and cuBLAS controls on each card and
the values and FP16 ratios listed at right.}
\caption{Achieved GB/s per arm on the L40S (filled) and the A100 (hollow),
joined; vertical rules mark FP16 and cuBLAS on each card. Ratios against
FP16 are formed on each card against its own control. Data:
\texttt{echelle-formats.csv}, \texttt{echelle-formats-a100.csv}.}
\label{fig:a100}
\end{figure}

A byte saved becomes time saved only while the kernel is bandwidth-bound,
so the scale of \S\ref{sec:scale} is a statement about a pair, format and
memory hierarchy. Nine of the arms were re-run from the same binary and
artifact on an NVIDIA A100 (HBM2e, \texttt{sm\_80}) against the L40S of
Table~\ref{tab:layouts} (GDDR6, \texttt{sm\_89}); the 2-bit competitor has
no A100 point. Kernels are compiled for the target at startup, nothing in
their text changed, and the f64 check returns the same worst errors on both
cards. Figure~\ref{fig:a100} joins each arm's achieved bandwidth on the two
cards.

\textbf{Every lattice arm falls below FP16 on the A100.} \textsc{Planes14}
runs at $0.79\times$ where it ran at $2.15\times$, and the mechanism is
visible in absolute terms. All lattice arms slow down on the card with more
bandwidth, \textsc{Planes14} from 428 to 250~GB/s, while FP16 converts the
HBM, 661 to 1052~GB/s, and cuBLAS 672 to 1204. The no-weights control goes
from $4.77\times$ to $1.68\times$ FP16, slowing by $\times 1.78$ in
absolute time (2.306 to 4.107~ms). SM clocks sampled once a second through
the bench show why: both cards run pinned at their maximum boost, 2\,520~MHz
on the L40S and 1\,410 on the A100, with no throttling and the only active
clock-event reason being the idle gap between kernels; these equal the
datasheet boost clocks~\citep{nvidia_l40s_datasheet,nvidia_a100_datasheet}. The
$\times 1.78$ slowdown is that $1.79$ clock ratio: a one-warp-per-row
geometry bound by instruction issue rather than by DRAM, its per-SM decode
throughput unchanged while the memory ceiling rises. This is clock evidence,
not the occupancy profile the platform's disabled counters would have given.
The 4-bit competitor stays above FP16, at $1.82\times$ from $3.38\times$.

What transfers is the ordering: \textsc{Planes14} leads, the two
\textsc{Golay70} decoders trail, and the hoisted decoder beats the first
on both cards. The
headline is bounded to the bandwidth-bound regime. On an HBM-class card,
expect Table~\ref{tab:layouts}'s ordering and none of its ratios.

\section{End-to-End Integration}
\label{sec:integration}

\textbf{Setup.} The served path is a decode-phase GEMV at batch size~1: a
prompt of $\ell$ tokens is processed as $\ell$ matrix--vector products (no
prefill GEMM), the activation rotation runs on CUDA only, and VRAM figures
count weights only (\S\ref{sec:limitations}). Decoding is greedy on a single
L40S, on Qwen3-4B unless another size is named. The fused \textsc{Planes14}
kernel replaces the 252 linear projections of \texttt{candle}~0.9.2
\citep{candle2023}, a Rust inference engine; a rotation kernel, verified
bit-exact on device, precedes each projection because the stored weights are
incoherence-rotated. The embedding table (9.7\% of the model, outside the
2-bit format) is int8 with per-group scales, measured free on both quality
metrics (\S\ref{sec:evaluation}). Correctness gate, on top of the row-level f64 verification of
\S\ref{sec:decoder}: the two engines emit identical greedy tokens for 88
positions, then part at token~89 on a tie-break. That divergence is
reproduced at the same position in every invocation, both continuations
remain fluent after it, and a 32-token run shows none.

\begin{table}[t]
\centering
\small
\begin{tabular}{lrrr}
\toprule
Phase (ms/token, median) & dense f16 & fused, f16 head & fused, q8 head \\
\midrule
embedding & 0.026 & 0.025 & 0.017 \\
transformer blocks & 13.291 & 10.439 & 10.432 \\
\texttt{lm\_head} & 26.672 & 25.886 & \textbf{0.598} \\
argmax + misc & 0.101 & 0.097 & 0.078 \\
\bottomrule
\end{tabular}
\caption{Per-token phase attribution (sync-bounded medians; phases
attribute time but their sum is not a throughput). Data:
\texttt{phases.csv}.}
\label{tab:phases}
\end{table}

\textbf{The output head.} The dense arm spends 26.7~ms per token in the
output head, twice the cost of its 36 transformer blocks
(Table~\ref{tab:phases}). The cause is a tensor primitive. The head applies
$hW^{\top}$ through the engine's broadcast matmul, whose rank-2
right-hand-side path materializes a transposed copy of the vocabulary table
on every token, roughly 0.8~GB at this vocabulary; the primitive's own
source marks the path as provisional. An int8 head kernel that reads its
413~MB in place brings the phase from 25.9 to 0.6~ms. The copy scales with
the vocabulary projection: 1.24~GB per token on the 8B (hidden size 4096,
untied head), where the same replacement doubles throughput (34.1 to
68.2~tok/s). The engine's \texttt{Linear} layer folds leading dimensions
into the row dimension and never takes this path, so its stock models do
not pay the copy. Our dense baseline does, because it calls the primitive
directly. The kernel statement below therefore holds the head identical in
both arms.

\textbf{Two formulations.} Table~\ref{tab:e2e} reports each size twice.
End to end, the served 4B engine reaches 87.0~tok/s against the dense
engine's 43.5, a $2.00\times$ gain. Most of it is the head replacement,
which the dense arm could adopt as well; the fenced phases of
Table~\ref{tab:phases} do not isolate its share, since synchronization
serializes work the real decode overlaps. The second formulation holds the
f16 head identical in both arms and measures the kernel plus format alone:
$1.11\times$ at 4B (48.3 against 43.5~tok/s), $1.29\times$ at 8B and
$1.41\times$ at 14B. That series grows monotonically with model size. The
served series (2.00, 2.57, 2.55) does not, because its denominator carries
a head path whose cost changes with the vocabulary projection. The 14B
is served only after raising the rotation kernel's shared-memory bound by
opt-in (\S\ref{sec:limitations}); its 128 greedy tokens match the dense
arm's, and so do the 8B's. Its memory ratio is direct, $3.14\times$ less
(9.39 against 29.54~GB, two host-side byte counts by the same runner), and
the same reading lands $0.38\%$ from the byte-exact whole-model rate of
\S\ref{sec:evaluation}.

\begin{table}[t]
\centering
\footnotesize
\setlength{\tabcolsep}{4pt}
\begin{tabular}{lrrrrrrr}
\toprule
 & & \multicolumn{2}{c}{same f16 head} & \multicolumn{2}{c}{served (int8 head)} & \multicolumn{2}{c}{VRAM (GB)} \\
\cmidrule(lr){3-4} \cmidrule(lr){5-6} \cmidrule(lr){7-8}
Size & dense tok/s & fused tok/s & gain & fused tok/s & gain & dense & served \\
\midrule
4B  & 43.5 [43.4--43.6] & 48.3 [48.1--48.3] & $\mathbf{1.11\times}$ & 87.0 [86.8--87.0] & $2.00\times$ & 8.04  & 2.60 \\
8B  & 26.5 [26.4--26.5] & 34.1 [34.0--34.1] & $\mathbf{1.29\times}$ & 68.2 [68.2--68.3] & $2.57\times$ & 16.38 & 5.45 \\
14B & 17.0 [17.0--17.0] & 23.9 [23.8--24.0] & $\mathbf{1.41\times}$ & 43.3 [43.2--43.4] & $2.55\times$ & 29.54 & 9.39 \\
\bottomrule
\end{tabular}
\caption{End-to-end decode on one L40S, 128 greedy tokens, three model
sizes. The bold same-head series is the kernel-and-format measurement; the
served series includes the output-head replacement both arms could adopt.
Medians of five timed generations with min--max ranges; ratios are
quotients of medians (arms never coexist). The dense arm is re-timed in each
invocation (medians within 0.1~tok/s); ratios form within one invocation. The 4B (2.60~GB) and 8B (5.45~GB) VRAM
cells are the nvidia-smi card reading from the quality-campaign
invocations; the 14B cell (9.39~GB) is a runner host byte count of weights
only, from the serving run. The byte-exact host figures are 2.595~GB for
the 4B (the 5.162~b/param of \S\ref{sec:evaluation}) and 5.41~GB for the
8B. Data: \texttt{campagne-finale.csv}, \texttt{tableau-8b.csv}, and the
serving runs' journal for the 14B row.}
\label{tab:e2e}
\end{table}

\textbf{What quantization buys inside its own stack.} We also timed the
official 4-bit checkpoint in its engine (vLLM~0.26.0, same L40S, prompt
token ids and 128-token protocol), with the f16 model as in-engine control.
Quantization buys $2.413\times$ there, $[2.412, 2.414]$, against
$1.11\times$ here; the two f16 controls differ (83.09 against our
43.5~tok/s), so the two within-stack ratios are reported side by side and
never divided.


\section{Evaluation}
\label{sec:evaluation}

\textbf{Protocol.} All quality numbers come from one harness on one machine
(NVIDIA L40S), f16 on both sides, with token-level fingerprints required to
match across arms. Perplexity is scored on WikiText-2
test~\citep{wikitext2016} at context 4096 over its first 12 non-overlapping
windows, 49\,140 scored tokens. The 12 windows are a prefix of the test
split. MMLU~\citep{mmlu2021} uses 2\,280 questions, 40 per subject from a
fixed seed, micro-averaged as in the source paper. The two fingerprints are
in Appendix~\ref{app:protocol}. The FP16 arm validates
the harness: $70.32 \pm 1.28$ on Qwen3-4B~\citep{qwen3_2025} against the 70.2
the source paper reports~\citep{llvq2026}. The quantized weights are the
sealed artifact: 47-bit indices plus one gain bit, from the error-propagating
GPTQ loop with shape--gain coding and input rotation, calibrated on 131k
tokens of C4~\citep{c4corpus}. The source paper calibrates on 6\,100
sequences of DCLM-edu, roughly two orders of magnitude more tokens.
Every score is taken on the written file. Table~\ref{tab:rates} collects
the five rates this paper uses with their denominators; no two are ever
subtracted or divided against each other.

\begin{table}[t]
\centering
\scriptsize
\setlength{\tabcolsep}{4pt}
\begin{tabular}{llrl}
\toprule
Rate & Denominator & Value (4B) & Includes \\
\midrule
Code & quantized weights & 2.000 & 47-bit index + 1 gain bit \\
Effective & quantized weights & 2.07 & + f32 tail columns, row scales \\
File & quantized weights & 2.17 & as written: headers, f64 centroids \\
Kernel (served) & streamed projection weights & 4.804 & unfolded payload + tail + scales (\S\ref{sec:layouts}) \\
Whole-model & all parameters & 5.162 & embedding and head included \\
\bottomrule
\end{tabular}
\caption{The five rates of this paper and their denominators; a comparison
is valid only within one row (values: the published 4B).}
\label{tab:rates}
\end{table}

\begin{table}[t]
\centering
\footnotesize
\begin{tabular}{lrrrr}
\toprule
 & FP16 & AWQ 4-bit & LLVQ (no kernel) & LLVQ fused (ours) \\
\midrule
disk & 8.04 GB & 2.67 GB & 1.77 GB & \textbf{1.41 GB}$^{\ddagger}$ \\
VRAM & 8.04 GB & 5.30 b/param$^{\dagger}$ & 8.04 GB & \textbf{2.60 GB / 5.162 b/param}$^{\S}$ \\
throughput & 43.5 [43.4--43.6] & ---$^{\dagger}$ & 43.5 [43.4--43.6] & \textbf{87.0 [86.8--87.0]} \\
ppl (WikiText-2) & 12.2369 & 13.5207 ($\times$1.105) & 16.9422 ($\times$1.384) & 16.9358 ($\times$1.384) \\
MMLU micro (\%) & 70.32 $\pm$ 1.28 & 70.04 $\pm$ 1.25 & 55.59 $\pm$ 1.35 & 55.70 $\pm$ 1.35 \\
\bottomrule
\end{tabular}
\caption{The 4B campaign: one harness, matching token fingerprints across
arms. $^{\dagger}$AWQ runs in its own engine (speed incomparable,
\S\ref{sec:integration}); its VRAM figure is that engine's accounting.
$^{\ddagger}$Quality scored on the 1.41~GB pre-baked int8-embedding file;
speed and VRAM measured from the 1.77~GB sealed file with the same
quantization applied at load, verified bit-identical against those bytes.
$^{\S}$VRAM provenance as in Table~\ref{tab:e2e}; the byte-exact host
figure is 2.595~GB (the 5.162~b/param above). Data:
\texttt{campagne-finale.csv}.}
\label{tab:campaign}
\end{table}

Columns 3 and 4 of Table~\ref{tab:campaign} serve the same quantized
linears. The transcoding is a proven bijection of decoded content on all
150\,681\,600 blocks, every output row is verified against f64 before any
timing (\S\ref{sec:decoder}), and both engines emit identical greedy tokens
up to a tie-break. The one lossy change on the served path is the int8
embedding, and the table shows it moving perplexity and MMLU within sampling
noise. On the whole-model row of Table~\ref{tab:rates} the served 4B
footprint is 5.162~b/param against 5.302 for the official AWQ checkpoint;
on the kernel row the same competitor reads less than we do (4.18 against
4.80).

\textbf{What 2 bits costs on this artifact.} Against FP16 the 4B pays $\times1.384$
perplexity and $-14.6$ MMLU points on the served int8 arm, $-14.73$ on the
sealed artifact; the two differ by less than either sampling error. The
sealed artifact carries every paired statistic below, because its
per-question dumps are the ones committed. Paired intervals are oriented
FP16 minus the quantized arm (positive is loss), from subject-stratified
paired bootstraps over the same 2\,280 questions: $+0.27$ $[-1.63, +2.13]$
for the 4-bit arm, containing zero, against $+14.73$ $[+11.98, +17.47]$ for
the 2-bit arm. The loss concentrates in reasoning-heavy subjects: abstract
algebra and professional accounting fall to 25\%, chance level, while
European history, international law and high-school psychology stay at or
above 80\%. On a 4B model, 4 bits wins on every axis but disk and served
footprint.

\begin{table}[t]
\centering
\footnotesize
\setlength{\tabcolsep}{3.5pt}
\begin{tabular}{lrrrr}
\toprule
 & FP16 & AWQ 4-bit & LLVQ fused, f16 heads & LLVQ fused, int8 heads (ours) \\
\midrule
disk & 16.38 GB & 6.10 GB & 4.32 GB & \textbf{3.16 GB} \\
VRAM & 16.38 GB & 6.10 GB / 5.96 b/param$^{\dagger}$ & 6.62 GB / 6.46 b/param & \textbf{5.45 GB / 5.32 b/param} \\
throughput & 26.5 [26.4--26.5] & ---$^{\dagger}$ & 34.1 [34.0--34.1] & \textbf{68.2 [68.2--68.3]} \\
ppl (WikiText-2) & 8.9899 & 9.4211 ($\times$1.048) & 10.9682 ($\times$1.220) & 10.9682 ($\times$1.220) \\
MMLU micro (\%) & 76.08 $\pm$ 1.21 & 73.01 $\pm$ 1.26 & 65.52 $\pm$ 1.31 & 65.63 $\pm$ 1.31 \\
\bottomrule
\end{tabular}
\caption{The 8B campaign: same harness, fingerprints, codebook and
calibration as the 4B, so that model size is the only variable. Column 3 is
the sealed artifact with f16 heads, the arm behind every paired statistic;
column 4 is the served configuration. Qwen3-8B unties its embedding from its
output head, so the served path stores both tables as int8, measured free:
perplexity identical to four decimals. $^{\dagger}$As in
Table~\ref{tab:campaign}; no vLLM timing exists for this checkpoint, which
pins no revision. Data: \texttt{tableau-8b.csv}.}
\label{tab:campaign8b}
\end{table}

\begin{figure}[t]
\centering
\includegraphics[width=\linewidth]{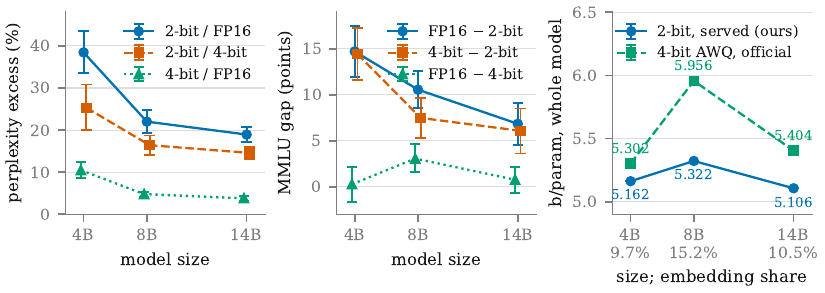}
\Description{Three panels against model size 4B, 8B and 14B: perplexity
excess in percent with paired confidence intervals for three arm pairs,
MMLU gap in points with paired intervals for three pairs, and whole-model
bits per parameter for the served configuration and the AWQ checkpoint.}
\caption{Three models, one configuration, one harness, matching token
fingerprints on all nine arms. Left: perplexity excess over the reference
with paired 95\% intervals over the 12 windows. Middle: MMLU gaps in points,
subject-stratified paired bootstraps over the same 2\,280 questions. Right:
whole-model b/param, embedding included, served configuration against the
official AWQ checkpoint, with the embedding share annotated. Cells in
Table~\ref{tab:scale}. Data: \texttt{ppl-appariee.csv},
\texttt{mmlu-appariee.csv}, \texttt{echelle-4b-8b.csv}.}
\label{fig:scale}
\end{figure}

\textbf{The scale point: three models.} Table~\ref{tab:campaign8b} is the
second model; a third, Qwen3-14B, was quantized under the same configuration
and scored in the same harness. Figure~\ref{fig:scale} sets the three side
by side, with the cells in Table~\ref{tab:scale} (Appendix~\ref{app:scale}).
The direction holds on both axes. Perplexity degradation falls
$\times1.3845 \to \times1.2201 \to \times1.1894$; the MMLU loss paired
against FP16 falls 14.73, 10.57, 6.85 points; the paired gap to the 4-bit
baseline falls 14.45, 7.49, 6.09 points, sealed-artifact arms throughout.
Read as excess over FP16, perplexity falls by 42.8\% $[-51.8, -33.5]$ on the
first step and 13.9\% $[-22.8, -4.9]$ on the second, the intervals from
pairing the 12 windows across sizes. Whether the second step is smaller than
the first is a question the intervals answer differently per metric and that
depends on the calibration draw (Appendix~\ref{app:scale}); we publish no
scaling law from three models.

\textbf{What one calibration draw is worth.} Three requantizations of the
4B differing only in the calibration draw (same corpus, configuration and
evaluation fingerprint) give perplexities of 16.74, 15.88 and 15.10, a range
of 10.3\% of the median. Every pairwise difference is separated from zero
window by window ($t = 4.5$, $10.9$, $7.7$). The published artifact (16.94)
sits above all three and is not a fourth draw of the same distribution: it
was calibrated on a contiguous prefix, read from a different C4 shard.
Absolute degradation figures at this calibration budget carry that variance;
comparisons at fixed artifact do not.

\textbf{Does the 4-bit baseline start paying?} In the same orientation the
4-bit loss against FP16 is $+3.07$ points at 8B, $[+1.61, +4.69]$, excluding
zero; at 4B ($+0.27$, above) and at 14B ($+0.76$, $[-0.65, +2.17]$) the
intervals contain zero. The finding is specific to the 8B, and no monotone
tendency is implied.

\textbf{Memory at three sizes.} Whole-model b/param, embedding included, is
below the official AWQ checkpoint at every size: 5.162, 5.322, 5.106 against
5.302, 5.956, 5.404, margins of 2.6\%, 10.6\% and 5.5\%
(Figure~\ref{fig:scale}, right). The margin is not monotone and carries no
trend: it tracks the embedding share (9.7\%, 15.2\%, 10.5\%), a table AWQ
leaves in FP16 and we store as int8. At 8B the trade is 7.49 MMLU points
against 10.6\% of VRAM and half the disk.

The served default is Planes14. Its sparse-overlay sibling Planes12x, wired
into the same runner, serves the 4B at 85.0~tok/s [84.7--85.1] in 2.36~GB,
a $\times1.96$ over the dense arm and $3.41\times$ less card memory, greedy
tokens identical to the dense arm through the same tie-break. It gives up
about 2\% of Planes14's throughput for a smaller footprint, the most compact
served configuration we measure.

\textbf{Published 2-bit results, respective protocols.} Table~\ref{tab:lit} places the
sealed arms beside the source paper's results at matched code rate,
48 bits per 24-weight block on every 2-bit row. At 4B our $\times1.38$ sits
level with QTIP and the 0-gain-bit LLVQ line ($\times1.37$) and behind the
2-gain-bit line ($\times1.25$); at 8B ($\times1.22$) we trail both LLVQ
lines and lead QTIP ($\times1.24$). The 4B MMLU shortfall (55.59 against
60.7, on a baseline our harness reproduces to 0.12 points) is wider than
the perplexity suggests. No difference carries an error bar: the
source paper publishes none, and question samples differ across papers.

\begin{table}[t]
\centering
\small
\setlength{\tabcolsep}{4.5pt}
\begin{tabular}{lcrrrrr}
\toprule
 & payload & \multicolumn{3}{c}{Qwen3-4B} & \multicolumn{2}{c}{Qwen3-8B} \\
\cmidrule(lr){3-5}\cmidrule(lr){6-7}
Method, no fine-tuning & b/weight & ppl & $\times$ & MMLU & ppl & $\times$ \\
\midrule
FP16 baseline, source paper & 16 & 12.41 & --- & 70.2 & 8.99 & --- \\
FP16 baseline, our harness & 16 & 12.2369 & --- & 70.32 & 8.9899 & --- \\
\midrule
QuIP\#/E8P12 & 2.000 & 21.15 & 1.70 & 48.6 & --- & --- \\
QTIP (3INST) & 2.000 & 17.04 & 1.37 & 57.4 & 11.17 & 1.24 \\
LLVQ shape--gain, 0 gain bits & 2.000 & 17.05 & 1.37 & 60.7 & 10.19 & 1.13 \\
LLVQ shape--gain, 2 gain bits & 2.000 & 15.54 & 1.25 & 59.3 & 10.82 & 1.20 \\
\midrule
Ours: $\Lambda_{24}(12)$, 1 gain bit & 2.000 & 16.9422 & 1.38 & 55.59 & 10.9682 & 1.22 \\
\bottomrule
\end{tabular}
\caption{Sealed-artifact arms beside the source paper's 2-bit results on
the same models (WikiText-2, context 4096, no fine-tuning). Rows 1 and 3--6
transcribed from its Table~6; rows 2 and 7 from
Tables~\ref{tab:campaign} and~\ref{tab:campaign8b}. Payload is the 2.000
code rate, 48 bits per 24-weight block; our effective rate is 2.07 (2.17 as
written in the file), see Protocol. $\times$ is against each harness's FP16 baseline; it publishes
no 8B MMLU and no error bars.}
\label{tab:lit}
\end{table}

\section{Limitations and Validity Envelope}
\label{sec:limitations}

\textbf{The validity envelope.} Table~\ref{tab:validity} states where the
measurements hold and where they stop. Two rows deserve a word: quality is
scored through the dense path, the served path's equivalence resting on the
bit-exact transcoding and row-level f64 checks of \S\ref{sec:decoder}; and
load-time unfolding is absorbed into a load that stays shorter than the
dense arm's: 122--131~s for the fused 4B, unfolding included, against
186--188~s to read 8~GB of f16.

\begin{table}[t]
\centering
\scriptsize
\setlength{\tabcolsep}{4pt}
\begin{tabular}{>{\raggedright\arraybackslash}p{0.16\linewidth}>{\raggedright\arraybackslash}p{0.42\linewidth}>{\raggedright\arraybackslash}p{0.35\linewidth}}
\toprule
Dimension & Measured domain & Boundary \\
\midrule
Phase, batch & decode-phase GEMV at batch 1; a prompt of $\ell$ tokens costs $\ell$ passes (five-token prompt end to end) & no prefill GEMM on the served path; time-to-first-token at long prompts unmeasured \\
Hardware & L40S: bandwidth-bound, byte savings convert to time & A100: every lattice arm below FP16 (\S\ref{sec:a100}) \\
Models & Qwen3 4B, 8B, 14B served end to end & 32B: rotation exceeds the shared-memory bound (Table~\ref{tab:envelope}) \\
Load & unfolding at load: 84~s (Planes14) and 404~s (Planes12x) on the 4B, 16 threads & unfolded stream not persisted; the cost recurs per process \\
Memory, platform & VRAM figures count weights only; rotation kernel is CUDA-only & KV cache excluded; served path NVIDIA-only \\
\bottomrule
\end{tabular}
\caption{The validity envelope: the domain each result was measured in, and
the boundary past which it is not claimed.}
\label{tab:validity}
\end{table}

\textbf{The MMLU deficit is unexplained.} The deficit of
\S\ref{sec:evaluation} is reproduced at three sizes. Three suspects are
bounded. Output-side rotation has a near-zero marginal effect
in the original paper's ablation. A free-magnitude variant with a
closed-form scale solve costs $\times1.99$ perplexity on a 0.6B proxy.
Calibration volume (131k tokens, about $100\times$ below the original
paper) is bounded on perplexity only: a contamination oracle recovers 1.6\%,
a $\times13$ volume curve 1.2\%. The leading untested suspect is
calibration composition, not volume: the source paper calibrates on
DCLM-edu, an education- and reasoning-curated corpus, where ours uses C4.
The perplexity excess reproduces the source ($\times1.38$ against its
$\times1.37$) while the MMLU shortfall does not, the signature a
reasoning-curated calibration would leave, since the loss concentrates in
reasoning-heavy subjects (\S\ref{sec:evaluation}). These data cannot
separate a faithful reproduction of a method that loses reasoning quality
at 2 bits from a bias introduced by C4 calibration; a source-matched
requantization would settle it. Learned per-column scales and low-rank
compensation remain untested.

\textbf{The trellis kernel is faster.} The 2-bit competitor's kernel runs
$2.27\times$ faster than Planes14 in one process (\S\ref{sec:qtip}). A
tuning pass on either side could move the number; none was attempted. The
same gap stands, unmeasured here, against Marlin's batched 4-bit GEMM and
the 2-bit vector-quantized systems of \S\ref{sec:related}.

\textbf{Batching.} The kernel is bandwidth-bound at batch~1 because one
activation amortizes each weight byte; a batched GEMM amortizes it across
rows, where the deployed 4-bit kernels already operate. Larger models change
the embedding share and the tail-column overhead, both of which enter the
whole-model rate.

\textbf{Three points, no law.} The direction holds on three sizes and
supports no scaling law (\S\ref{sec:evaluation}, Appendix~\ref{app:scale}).
The 70B class, the regime the memory argument targets, is measurable with
this methodology and not measured here.

\textbf{The rotated basis and the shared-memory wall.} Weights are stored
in the incoherence-rotated basis, so every consumer applies the activation
rotation, a Walsh--Hadamard transform whose barrier ladder synchronizes one
thread block. That block must hold the whole activation in shared memory,
one f32 per coordinate, and the binding width is the \texttt{down\_proj}
input (Table~\ref{tab:envelope}). The 14B required the opt-in 101\,376-byte
bound, measured on the serving card. The 32B needs 102\,400 bytes and
misses that bound by 1\,024 bytes, which no host-side trimming reclaims;
every layout of \S\ref{sec:layouts} calls the same rotation.
Half-precision shared activations would halve the footprint, at an
accumulation depth of order $\log_2 n \approx 14.6$ stages at
$n = 25\,600$ that we have not priced.

\begin{table}[t]
\centering
\footnotesize
\setlength{\tabcolsep}{4pt}%
\begin{tabular}{lrrll}
\toprule
Model & \texttt{intermediate\_size} & shared bytes & vs.\ 49\,152 default & vs.\ 101\,376 opt-in \\
\midrule
Qwen3-4B  &  9\,728 &  38\,912 & fits & fits \\
Qwen3-8B  & 12\,288 &  49\,152 & fits exactly & fits \\
Qwen3-14B & 17\,408 &  69\,632 & does not fit & fits \\
Qwen3-32B & 25\,600 & 102\,400 & does not fit & short by 1\,024 bytes \\
\bottomrule
\end{tabular}
\caption{Shared memory required by the Walsh--Hadamard rotation of the
\texttt{down\_proj} input (one f32 per coordinate) against the default and
opt-in per-block bounds measured on the serving card.}
\label{tab:envelope}
\end{table}

\section{Related Work}
\label{sec:related}

\textbf{Extreme low-bit weight quantization.} GPTQ~\citep{gptq2023}
established the error-propagating framework most post-training methods
refine. At 2 bits, vector and trellis codes replaced scalar grids:
QuIP~\citep{quip2023} introduced incoherence processing,
QuIP\#~\citep{quipsharp2024} added E8-lattice codebooks, and
QTIP~\citep{qtip2024} reached the previous state of the art with trellis
codes. LLVQ~\citep{llvq2026} improved on QTIP's quality with a spherical GPTQ
variant on the Leech lattice. We implement its quantizer independently and
supply the decoder it leaves open: its kernel decodes one shell. QTIP's
kernel is timed in our harness (\S\ref{sec:qtip}). Lattice codebooks that
fit a lookup table are the counterpart of ours. The E8P codebook of QuIP\#
holds $2^{16}$ points; the IQ2\_XXS, IQ2\_XS and IQ2\_S formats of
llama.cpp~\citep{llamacpp}, software without a publication, use E8-derived
grids of 256 to 1\,024 points with fused kernels. Those codebooks decode by
lookup; $\Lambda_{24}(12)$ holds $1.1 \times 10^{14}$ points and must be
unfolded (\S\ref{sec:decoder}). The lattice and its Golay construction are
classical~\citep{conway1999sphere,cohn2017}; Leech nearest-neighbor search
dates to \citet{adoul1988}.

SpQR~\citep{spqr2023} and SqueezeLLM~\citep{squeezellm2024} keep a scalar
grid and isolate outlier weights into a sparse side-channel, the move our
sparse overlay makes (\S\ref{sec:layouts}) at the same price: a second
irregular memory stream. AQLM~\citep{aqlm2024} and VPTQ~\citep{vptq2024} run
at our rate with released kernels; GPTVQ~\citep{gptvq2024}, from the group
behind LLVQ, precedes that line. Among the released implementations of
these systems we found none serving the multi-shell $\Lambda_{24}$
codebook. We measure none of them
(\S\ref{sec:limitations}). Both are lookup-decoded, so they sit on the
fits-a-lookup-table side of the frontier above that the E8P and IQ2 entry
counts already mark; $\Lambda_{24}(12)$'s $1.1\times10^{14}$ points are the
sole must-unfold member, and timing them would confirm that side rather than
test the unfolding cost.

\textbf{Rotations.} QuaRot~\citep{quarot2024} and SpinQuant~\citep{spinquant2024}
made incoherence rotations practical at 4 bits; each applies a serving-time
Hadamard transform and meets the shared-memory constraint of
\S\ref{sec:limitations} at some width, none removes it.

\textbf{Kernels for quantized inference.} Below 8 bits the \emph{decode} sets
the cost of a GEMV, the premise of \S\ref{sec:layouts}.
LUT-GEMM~\citep{lutgemm2024} and FLUTE~\citep{flute2024} answer with lookup
tables sized for fast memory, as our class table does; a Leech class is a
permutation pattern rather than a value, so the table is read once per block
rather than per weight. AWQ~\citep{awq2024} is the deployed 4-bit reference
we port unchanged into our harness; Marlin~\citep{marlin2024} shows what a
mature mixed-precision kernel achieves at 4 bits, including batched regimes
outside our scope. ExLlamaV2~\citep{exllamav2} ships comparable formats
without a publication; we cite it as software. We ask instead how far below
4 bits the \emph{in-VRAM} rate can go before decode cost overtakes byte
savings (\S\ref{sec:scale}).

\textbf{Performance models.} The quantity organizing \S\ref{sec:layouts}, the
fraction of its byte bound each layout converts into time, is a roofline
argument~\citep{roofline2009} with two adaptations. First, the byte bound is
the FP16 control's measured bandwidth, and a no-weights control prices the
launch geometry. Second, the same layouts run on two memory hierarchies
(\S\ref{sec:a100}): on an L40S byte savings convert into time; on an A100
the ordering survives and the ratios do not. One machine's roofline does not
describe the other.

\textbf{Serving engines.} Our integration builds on
\texttt{candle}~\citep{candle2023}; the 4-bit throughput reference of
\S\ref{sec:integration} was measured in vLLM~\citep{vllm2023}, a different
engine. The output-head materialization of \S\ref{sec:integration} belongs to
the engine's broadcast matmul primitive; \texttt{candle}'s served models
reach their heads through \texttt{Linear} and avoid it.

\section{Conclusion}
\label{sec:conclusion}

We wrote a fused dequantize-plus-matvec kernel for the full 301-class
$\Lambda_{24}(12)$ codebook, to our knowledge the first: divergence-free, verified against
f64, 65\% of its byte bound on an L40S. The in-VRAM rate is a distinct design
axis: four bit-exact layouts span 5.51 to 3.59~b/weight, and bit planes beat
one-hot masks on size and speed at constant bandwidth. The curve turns
between 4.8 and 4.3~b/weight, where a second, irregular stream (the
exception side channel) enters, and collapses at 3.6, where the decode stops
being shifts and masks. Against deployed 4-bit and 2-bit GEMV kernels
in one process, the trellis kernel reads 2.40$\times$ fewer bytes and runs
2.27$\times$ faster at near-equal efficiency: the unfolding of a codebook
no lookup table holds is paid in bytes. The validity envelope is measured: the
trellis kernel outruns our no-weights control, launch geometry holds 39\% of
the gap to the DRAM floor, and fusing projections recovers 11.7\% of the
\textsc{Planes14} kernel time and $1.061\times$ end to end. On an A100 every lattice arm falls below
FP16, and same-head serving gains are 1.11$\times$, 1.29$\times$ and
1.41$\times$.

Two questions stay open: does the quality curve, whose direction holds on
three sizes without a scaling law, reach the 70B class, and which untested
lever (calibration composition, learned column scales, low-rank
compensation) closes the reasoning-concentrated MMLU deficit? A successor
should attack launch geometry and unfolding cost, not the factor against
FP16.

\section{Availability}
\label{sec:availability}

Code, the CSVs behind every table and figure, and the paper source are at
\ifanon{\url{https://github.com/pjmalandrino/llvq}}{a public repository whose
URL is withheld for double-anonymous review} (MIT or Apache-2.0);
\texttt{make} in \texttt{paper/} rejects a table that drifts from its CSV.
The measurement campaign cost under \$100 of rented GPU time in total;
per-job amounts, dates and logs are in the repository. The 2-bit
competitor's kernel is GPL~v3: the benchmark fetches it at a pinned commit
rather than redistributing it.

The quantized Qwen3-4B artifact (sealed, 1.77~GB, and with the int8 embedding
pre-baked, 1.41~GB; Table~\ref{tab:campaign}) is at
\ifanon{\url{https://huggingface.co/Pier-Jean/Qwen3-4B-LLVQ-2bit}}{a public
model repository whose URL is withheld for double-anonymous review}. The 8B
and 14B artifacts are unpublished; a requantization would be a new
calibration draw (\S\ref{sec:evaluation}) rather than the same object. The
MMLU dumps and per-window log-likelihoods behind every paired statistic are
committed raw. Serving needs the CUDA-only rotation kernel of
\S\ref{sec:limitations}.

Artifacts carry tokenizer and configuration; on an NVIDIA GPU the served
configuration and both evaluations replay from the file alone:

{\small
\begin{verbatim}
LLVQ_FUSED_LAYOUT=planes14 LLVQ_EMBED=q8 cargo run --release \
  -p llvq-llm --features cuda --bin fusedrun -- <artifact>

LLVQ_DTYPE=f16 cargo run --release \
  -p llvq-llm --features cuda --bin ppl -- 4096 12 cuda <artifact>

cargo run --release \
  -p llvq-llm --features cuda --bin mmlu -- <artifact> cuda 40
\end{verbatim}
}

\noindent
\texttt{LLVQ\_FUSED\_LAYOUT} selects the layout of \S\ref{sec:layouts}
(\texttt{planes14} default; anything unrecognised is an error),
\texttt{LLVQ\_EMBED} the embedding precision, \texttt{LLVQ\_DTYPE} the
evaluation dtype. Each binary prints a fingerprint of the scored tokens;
numbers compare only when fingerprints match.

\section*{Use of generative AI}

Disclosed under ACM's Policy on Authorship. Anthropic's Claude, through the
Claude Code CLI in 2026, drafted and revised this manuscript's text, wrote
the figure-generating and table-checking scripts, wrote portions of the CUDA
kernels, Rust host code and tests, and recomputed statistics from the
committed measurement journals. The author designed the study, authorized
every measurement, verified each number against its source, and is
responsible for the entire content.

\bibliographystyle{ACM-Reference-Format}
\bibliography{refs}


\begin{thebibliography}{30}


\ifx \showCODEN    \undefined \def \showCODEN     #1{\unskip}     \fi
\ifx \showISBNx    \undefined \def \showISBNx     #1{\unskip}     \fi
\ifx \showISBNxiii \undefined \def \showISBNxiii  #1{\unskip}     \fi
\ifx \showISSN     \undefined \def \showISSN      #1{\unskip}     \fi
\ifx \showLCCN     \undefined \def \showLCCN      #1{\unskip}     \fi
\ifx \shownote     \undefined \def \shownote      #1{#1}          \fi
\ifx \showarticletitle \undefined \def \showarticletitle #1{#1}   \fi
\ifx \showURL      \undefined \def \showURL       {\relax}        \fi
\providecommand\bibfield[2]{#2}
\providecommand\bibinfo[2]{#2}
\providecommand\natexlab[1]{#1}
\providecommand\showeprint[2][]{arXiv:#2}

\bibitem[Adoul and Barth(1988)]%
        {adoul1988}
\bibfield{author}{\bibinfo{person}{Jean-Pierre Adoul} {and}
  \bibinfo{person}{Michel Barth}.} \bibinfo{year}{1988}\natexlab{}.
\newblock \showarticletitle{Nearest neighbor algorithm for spherical codes from
  the {Leech} lattice}.
\newblock \bibinfo{journal}{\emph{IEEE Transactions on Information Theory}}
  \bibinfo{volume}{34}, \bibinfo{number}{5} (\bibinfo{year}{1988}),
  \bibinfo{pages}{1188--1202}.
\newblock


\bibitem[Ashkboos et~al\mbox{.}(2024)]%
        {quarot2024}
\bibfield{author}{\bibinfo{person}{Saleh Ashkboos}, \bibinfo{person}{Amirkeivan
  Mohtashami}, \bibinfo{person}{Maximilian~L. Croci}, \bibinfo{person}{Bo Li},
  \bibinfo{person}{Pashmina Cameron}, \bibinfo{person}{Martin Jaggi},
  \bibinfo{person}{Dan Alistarh}, \bibinfo{person}{Torsten Hoefler}, {and}
  \bibinfo{person}{James Hensman}.} \bibinfo{year}{2024}\natexlab{}.
\newblock \showarticletitle{{QuaRot}: Outlier-Free 4-Bit Inference in Rotated
  {LLMs}}. In \bibinfo{booktitle}{\emph{Advances in Neural Information
  Processing Systems (NeurIPS)}}.
\newblock
\newblock
\shownote{arXiv:2404.00456}.


\bibitem[Chee et~al\mbox{.}(2023)]%
        {quip2023}
\bibfield{author}{\bibinfo{person}{Jerry Chee}, \bibinfo{person}{Yaohui Cai},
  \bibinfo{person}{Volodymyr Kuleshov}, {and} \bibinfo{person}{Christopher
  De~Sa}.} \bibinfo{year}{2023}\natexlab{}.
\newblock \showarticletitle{{QuIP}: 2-Bit Quantization of Large Language Models
  With Guarantees}. In \bibinfo{booktitle}{\emph{Advances in Neural Information
  Processing Systems (NeurIPS)}}.
\newblock
\newblock
\shownote{arXiv:2307.13304}.


\bibitem[Cohn et~al\mbox{.}(2017)]%
        {cohn2017}
\bibfield{author}{\bibinfo{person}{Henry Cohn}, \bibinfo{person}{Abhinav
  Kumar}, \bibinfo{person}{Stephen~D. Miller}, \bibinfo{person}{Danylo
  Radchenko}, {and} \bibinfo{person}{Maryna Viazovska}.}
  \bibinfo{year}{2017}\natexlab{}.
\newblock \showarticletitle{The sphere packing problem in dimension 24}.
\newblock \bibinfo{journal}{\emph{Annals of Mathematics}}
  \bibinfo{volume}{185}, \bibinfo{number}{3} (\bibinfo{year}{2017}),
  \bibinfo{pages}{1017--1033}.
\newblock


\bibitem[Conway and Sloane(1999)]%
        {conway1999sphere}
\bibfield{author}{\bibinfo{person}{John~H. Conway} {and} \bibinfo{person}{Neil
  J.~A. Sloane}.} \bibinfo{year}{1999}\natexlab{}.
\newblock \bibinfo{booktitle}{\emph{Sphere Packings, Lattices and Groups}
  (\bibinfo{edition}{3rd} ed.)}.
\newblock \bibinfo{publisher}{Springer}.
\newblock


\bibitem[Dettmers et~al\mbox{.}(2024)]%
        {spqr2023}
\bibfield{author}{\bibinfo{person}{Tim Dettmers}, \bibinfo{person}{Ruslan
  Svirschevski}, \bibinfo{person}{Vage Egiazarian}, \bibinfo{person}{Denis
  Kuznedelev}, \bibinfo{person}{Elias Frantar}, \bibinfo{person}{Saleh
  Ashkboos}, \bibinfo{person}{Alexander Borzunov}, \bibinfo{person}{Torsten
  Hoefler}, {and} \bibinfo{person}{Dan Alistarh}.}
  \bibinfo{year}{2024}\natexlab{}.
\newblock \showarticletitle{{SpQR}: A Sparse-Quantized Representation for
  Near-Lossless {LLM} Weight Compression}. In
  \bibinfo{booktitle}{\emph{International Conference on Learning
  Representations (ICLR)}}.
\newblock
\newblock
\shownote{arXiv:2306.03078}.


\bibitem[Egiazarian et~al\mbox{.}(2024)]%
        {aqlm2024}
\bibfield{author}{\bibinfo{person}{Vage Egiazarian}, \bibinfo{person}{Andrei
  Panferov}, \bibinfo{person}{Denis Kuznedelev}, \bibinfo{person}{Elias
  Frantar}, \bibinfo{person}{Artem Babenko}, {and} \bibinfo{person}{Dan
  Alistarh}.} \bibinfo{year}{2024}\natexlab{}.
\newblock \showarticletitle{Extreme Compression of Large Language Models via
  Additive Quantization}. In \bibinfo{booktitle}{\emph{Proceedings of the 41st
  International Conference on Machine Learning (ICML)}}.
  \bibinfo{pages}{12284--12303}.
\newblock
\newblock
\shownote{arXiv:2401.06118}.


\bibitem[Frantar et~al\mbox{.}(2023)]%
        {gptq2023}
\bibfield{author}{\bibinfo{person}{Elias Frantar}, \bibinfo{person}{Saleh
  Ashkboos}, \bibinfo{person}{Torsten Hoefler}, {and} \bibinfo{person}{Dan
  Alistarh}.} \bibinfo{year}{2023}\natexlab{}.
\newblock \showarticletitle{{GPTQ}: Accurate Post-Training Quantization for
  Generative Pre-trained Transformers}. In
  \bibinfo{booktitle}{\emph{International Conference on Learning
  Representations (ICLR)}}.
\newblock
\newblock
\shownote{arXiv:2210.17323}.


\bibitem[Frantar et~al\mbox{.}(2025)]%
        {marlin2024}
\bibfield{author}{\bibinfo{person}{Elias Frantar}, \bibinfo{person}{Roberto~L.
  Castro}, \bibinfo{person}{Jiale Chen}, \bibinfo{person}{Torsten Hoefler},
  {and} \bibinfo{person}{Dan Alistarh}.} \bibinfo{year}{2025}\natexlab{}.
\newblock \showarticletitle{{MARLIN}: Mixed-Precision Auto-Regressive Parallel
  Inference on Large Language Models}. In \bibinfo{booktitle}{\emph{Proceedings
  of the 30th {ACM} {SIGPLAN} Annual Symposium on Principles and Practice of
  Parallel Programming (PPoPP)}}. \bibinfo{pages}{239--251}.
\newblock
\newblock
\shownote{arXiv:2408.11743}.


\bibitem[{ggml-org}(2023)]%
        {llamacpp}
\bibfield{author}{\bibinfo{person}{{ggml-org}}.}
  \bibinfo{year}{2023}\natexlab{}.
\newblock \bibinfo{title}{llama.cpp: {LLM} inference in {C/C++}}.
\newblock \bibinfo{howpublished}{\url{https://github.com/ggml-org/llama.cpp}}.
\newblock
\newblock
\shownote{Release v0.2.0, accessed 2026-08-22}.


\bibitem[Guo et~al\mbox{.}(2024)]%
        {flute2024}
\bibfield{author}{\bibinfo{person}{Han Guo}, \bibinfo{person}{William Brandon},
  \bibinfo{person}{Radostin Cholakov}, \bibinfo{person}{Jonathan Ragan-Kelley},
  \bibinfo{person}{Eric~P. Xing}, {and} \bibinfo{person}{Yoon Kim}.}
  \bibinfo{year}{2024}\natexlab{}.
\newblock \showarticletitle{Fast Matrix Multiplications for Lookup
  Table-Quantized {LLMs}}. In \bibinfo{booktitle}{\emph{Findings of the
  Association for Computational Linguistics: {EMNLP} 2024}}.
  \bibinfo{pages}{12419--12433}.
\newblock
\newblock
\shownote{arXiv:2407.10960}.


\bibitem[Hendrycks et~al\mbox{.}(2021)]%
        {mmlu2021}
\bibfield{author}{\bibinfo{person}{Dan Hendrycks}, \bibinfo{person}{Collin
  Burns}, \bibinfo{person}{Steven Basart}, \bibinfo{person}{Andy Zou},
  \bibinfo{person}{Mantas Mazeika}, \bibinfo{person}{Dawn Song}, {and}
  \bibinfo{person}{Jacob Steinhardt}.} \bibinfo{year}{2021}\natexlab{}.
\newblock \showarticletitle{Measuring Massive Multitask Language
  Understanding}. In \bibinfo{booktitle}{\emph{International Conference on
  Learning Representations (ICLR)}}.
\newblock
\newblock
\shownote{arXiv:2009.03300}.


\bibitem[{Hugging Face}(2023)]%
        {candle2023}
\bibfield{author}{\bibinfo{person}{{Hugging Face}}.}
  \bibinfo{year}{2023}\natexlab{}.
\newblock \bibinfo{title}{candle: Minimalist {ML} framework for {Rust}}.
\newblock \bibinfo{howpublished}{\url{https://github.com/huggingface/candle}}.
\newblock
\newblock
\shownote{Version 0.9.2}.


\bibitem[Kim et~al\mbox{.}(2024)]%
        {squeezellm2024}
\bibfield{author}{\bibinfo{person}{Sehoon Kim}, \bibinfo{person}{Coleman
  Richard~Charles Hooper}, \bibinfo{person}{Amir Gholami},
  \bibinfo{person}{Zhen Dong}, \bibinfo{person}{Xiuyu Li},
  \bibinfo{person}{Sheng Shen}, \bibinfo{person}{Michael~W. Mahoney}, {and}
  \bibinfo{person}{Kurt Keutzer}.} \bibinfo{year}{2024}\natexlab{}.
\newblock \showarticletitle{{SqueezeLLM}: Dense-and-Sparse Quantization}. In
  \bibinfo{booktitle}{\emph{Proceedings of the 41st International Conference on
  Machine Learning (ICML)}}. \bibinfo{pages}{23901--23923}.
\newblock
\newblock
\shownote{arXiv:2306.07629}.


\bibitem[Kwon et~al\mbox{.}(2023)]%
        {vllm2023}
\bibfield{author}{\bibinfo{person}{Woosuk Kwon}, \bibinfo{person}{Zhuohan Li},
  \bibinfo{person}{Siyuan Zhuang}, \bibinfo{person}{Ying Sheng},
  \bibinfo{person}{Lianmin Zheng}, \bibinfo{person}{Cody~Hao Yu},
  \bibinfo{person}{Joseph~E. Gonzalez}, \bibinfo{person}{Hao Zhang}, {and}
  \bibinfo{person}{Ion Stoica}.} \bibinfo{year}{2023}\natexlab{}.
\newblock \showarticletitle{Efficient Memory Management for Large Language
  Model Serving with {PagedAttention}}. In
  \bibinfo{booktitle}{\emph{Proceedings of the 29th Symposium on Operating
  Systems Principles (SOSP)}}. \bibinfo{pages}{611--626}.
\newblock
\newblock
\shownote{arXiv:2309.06180}.


\bibitem[Lin et~al\mbox{.}(2024)]%
        {awq2024}
\bibfield{author}{\bibinfo{person}{Ji Lin}, \bibinfo{person}{Jiaming Tang},
  \bibinfo{person}{Haotian Tang}, \bibinfo{person}{Shang Yang},
  \bibinfo{person}{Wei-Ming Chen}, \bibinfo{person}{Wei-Chen Wang},
  \bibinfo{person}{Guangxuan Xiao}, \bibinfo{person}{Xingyu Dang},
  \bibinfo{person}{Chuang Gan}, {and} \bibinfo{person}{Song Han}.}
  \bibinfo{year}{2024}\natexlab{}.
\newblock \showarticletitle{{AWQ}: Activation-aware Weight Quantization for
  On-Device {LLM} Compression and Acceleration}. In
  \bibinfo{booktitle}{\emph{Proceedings of Machine Learning and Systems
  (MLSys)}}, Vol.~\bibinfo{volume}{6}. \bibinfo{pages}{87--100}.
\newblock
\newblock
\shownote{arXiv:2306.00978}.


\bibitem[Liu et~al\mbox{.}(2024)]%
        {vptq2024}
\bibfield{author}{\bibinfo{person}{Yifei Liu}, \bibinfo{person}{Jicheng Wen},
  \bibinfo{person}{Yang Wang}, \bibinfo{person}{Shengyu Ye},
  \bibinfo{person}{Li~Lyna Zhang}, \bibinfo{person}{Ting Cao},
  \bibinfo{person}{Cheng Li}, {and} \bibinfo{person}{Mao Yang}.}
  \bibinfo{year}{2024}\natexlab{}.
\newblock \showarticletitle{{VPTQ}: Extreme Low-bit Vector Post-Training
  Quantization for Large Language Models}. In
  \bibinfo{booktitle}{\emph{Proceedings of the 2024 Conference on Empirical
  Methods in Natural Language Processing (EMNLP)}}.
  \bibinfo{pages}{8181--8196}.
\newblock
\newblock
\shownote{arXiv:2409.17066}.


\bibitem[Liu et~al\mbox{.}(2025)]%
        {spinquant2024}
\bibfield{author}{\bibinfo{person}{Zechun Liu}, \bibinfo{person}{Changsheng
  Zhao}, \bibinfo{person}{Igor Fedorov}, \bibinfo{person}{Bilge Soran},
  \bibinfo{person}{Dhruv Choudhary}, \bibinfo{person}{Raghuraman
  Krishnamoorthi}, \bibinfo{person}{Vikas Chandra}, \bibinfo{person}{Yuandong
  Tian}, {and} \bibinfo{person}{Tijmen Blankevoort}.}
  \bibinfo{year}{2025}\natexlab{}.
\newblock \showarticletitle{{SpinQuant}: {LLM} Quantization with Learned
  Rotations}. In \bibinfo{booktitle}{\emph{International Conference on Learning
  Representations (ICLR)}}.
\newblock
\newblock
\shownote{arXiv:2405.16406}.


\bibitem[Merity et~al\mbox{.}(2017)]%
        {wikitext2016}
\bibfield{author}{\bibinfo{person}{Stephen Merity}, \bibinfo{person}{Caiming
  Xiong}, \bibinfo{person}{James Bradbury}, {and} \bibinfo{person}{Richard
  Socher}.} \bibinfo{year}{2017}\natexlab{}.
\newblock \showarticletitle{Pointer Sentinel Mixture Models}. In
  \bibinfo{booktitle}{\emph{International Conference on Learning
  Representations (ICLR)}}.
\newblock
\newblock
\shownote{arXiv:1609.07843}.


\bibitem[{NVIDIA Corporation}(2020)]%
        {nvidia_a100_datasheet}
\bibfield{author}{\bibinfo{person}{{NVIDIA Corporation}}.}
  \bibinfo{year}{2020}\natexlab{}.
\newblock \bibinfo{title}{{NVIDIA A100} {Tensor Core} {GPU} datasheet}.
\newblock \bibinfo{howpublished}{NVIDIA Corporation}.
\newblock
\newblock
\shownote{Accessed 2026-08-22}.


\bibitem[{NVIDIA Corporation}(2023)]%
        {nvidia_l40s_datasheet}
\bibfield{author}{\bibinfo{person}{{NVIDIA Corporation}}.}
  \bibinfo{year}{2023}\natexlab{}.
\newblock \bibinfo{title}{{NVIDIA L40S} datasheet}.
\newblock \bibinfo{howpublished}{NVIDIA Corporation}.
\newblock
\newblock
\shownote{Accessed 2026-08-22}.


\bibitem[Park et~al\mbox{.}(2024)]%
        {lutgemm2024}
\bibfield{author}{\bibinfo{person}{Gunho Park}, \bibinfo{person}{Baeseong
  Park}, \bibinfo{person}{Minsub Kim}, \bibinfo{person}{Sungjae Lee},
  \bibinfo{person}{Jeonghoon Kim}, \bibinfo{person}{Beomseok Kwon},
  \bibinfo{person}{Se~Jung Kwon}, \bibinfo{person}{Byeongwook Kim},
  \bibinfo{person}{Youngjoo Lee}, {and} \bibinfo{person}{Dongsoo Lee}.}
  \bibinfo{year}{2024}\natexlab{}.
\newblock \showarticletitle{{LUT-GEMM}: Quantized Matrix Multiplication based
  on {LUTs} for Efficient Inference in Large-Scale Generative Language Models}.
  In \bibinfo{booktitle}{\emph{International Conference on Learning
  Representations (ICLR)}}.
\newblock
\newblock
\shownote{arXiv:2206.09557}.


\bibitem[{Qwen Team}(2025)]%
        {qwen3_2025}
\bibfield{author}{\bibinfo{person}{{Qwen Team}}.}
  \bibinfo{year}{2025}\natexlab{}.
\newblock \showarticletitle{Qwen3 Technical Report}.
\newblock \bibinfo{journal}{\emph{arXiv preprint arXiv:2505.09388}}
  (\bibinfo{year}{2025}).
\newblock


\bibitem[Raffel et~al\mbox{.}(2020)]%
        {c4corpus}
\bibfield{author}{\bibinfo{person}{Colin Raffel}, \bibinfo{person}{Noam
  Shazeer}, \bibinfo{person}{Adam Roberts}, \bibinfo{person}{Katherine Lee},
  \bibinfo{person}{Sharan Narang}, \bibinfo{person}{Michael Matena},
  \bibinfo{person}{Yanqi Zhou}, \bibinfo{person}{Wei Li}, {and}
  \bibinfo{person}{Peter~J. Liu}.} \bibinfo{year}{2020}\natexlab{}.
\newblock \showarticletitle{Exploring the Limits of Transfer Learning with a
  Unified Text-to-Text Transformer}.
\newblock \bibinfo{journal}{\emph{Journal of Machine Learning Research}}
  \bibinfo{volume}{21}, \bibinfo{number}{140} (\bibinfo{year}{2020}),
  \bibinfo{pages}{1--67}.
\newblock
\newblock
\shownote{arXiv:1910.10683}.


\bibitem[Tseng et~al\mbox{.}(2024a)]%
        {quipsharp2024}
\bibfield{author}{\bibinfo{person}{Albert Tseng}, \bibinfo{person}{Jerry Chee},
  \bibinfo{person}{Qingyao Sun}, \bibinfo{person}{Volodymyr Kuleshov}, {and}
  \bibinfo{person}{Christopher De~Sa}.} \bibinfo{year}{2024}\natexlab{a}.
\newblock \showarticletitle{{QuIP\#}: Even Better {LLM} Quantization with
  {Hadamard} Incoherence and Lattice Codebooks}. In
  \bibinfo{booktitle}{\emph{Proceedings of the 41st International Conference on
  Machine Learning (ICML)}}.
\newblock
\newblock
\shownote{arXiv:2402.04396}.


\bibitem[Tseng et~al\mbox{.}(2024b)]%
        {qtip2024}
\bibfield{author}{\bibinfo{person}{Albert Tseng}, \bibinfo{person}{Qingyao
  Sun}, \bibinfo{person}{David Hou}, {and} \bibinfo{person}{Christopher
  De~Sa}.} \bibinfo{year}{2024}\natexlab{b}.
\newblock \showarticletitle{{QTIP}: Quantization with Trellises and Incoherence
  Processing}. In \bibinfo{booktitle}{\emph{Advances in Neural Information
  Processing Systems (NeurIPS)}}.
\newblock
\newblock
\shownote{arXiv:2406.11235}.


\bibitem[{turboderp-org}(2023)]%
        {exllamav2}
\bibfield{author}{\bibinfo{person}{{turboderp-org}}.}
  \bibinfo{year}{2023}\natexlab{}.
\newblock \bibinfo{title}{{ExLlamaV2}: A fast inference library for running
  {LLMs} locally on modern consumer-class {GPUs}}.
\newblock
  \bibinfo{howpublished}{\url{https://github.com/turboderp-org/exllamav2}}.
\newblock
\newblock
\shownote{Version 0.3.2, accessed 2026-08-19}.


\bibitem[van Baalen et~al\mbox{.}(2024)]%
        {gptvq2024}
\bibfield{author}{\bibinfo{person}{Mart van Baalen}, \bibinfo{person}{Andrey
  Kuzmin}, \bibinfo{person}{Ivan Koryakovskiy}, \bibinfo{person}{Cedric
  Bastoul}, \bibinfo{person}{Peter Couperus}, \bibinfo{person}{Eric Mahurin},
  \bibinfo{person}{Tijmen Blankevoort}, \bibinfo{person}{Markus Nagel}, {and}
  \bibinfo{person}{Paul Whatmough}.} \bibinfo{year}{2024}\natexlab{}.
\newblock \showarticletitle{{GPTVQ}: The Blessing of Dimensionality for {LLM}
  Quantization}.
\newblock \bibinfo{journal}{\emph{arXiv preprint arXiv:2402.15319}}
  (\bibinfo{year}{2024}).
\newblock


\bibitem[van~der Ouderaa et~al\mbox{.}(2026)]%
        {llvq2026}
\bibfield{author}{\bibinfo{person}{Tycho F.~A. van~der Ouderaa},
  \bibinfo{person}{Mart van Baalen}, \bibinfo{person}{Paul Whatmough}, {and}
  \bibinfo{person}{Markus Nagel}.} \bibinfo{year}{2026}\natexlab{}.
\newblock \showarticletitle{Leech Lattice Vector Quantization for Efficient
  {LLM} Compression}.
\newblock \bibinfo{journal}{\emph{arXiv preprint arXiv:2603.11021}}
  (\bibinfo{year}{2026}).
\newblock


\bibitem[Williams et~al\mbox{.}(2009)]%
        {roofline2009}
\bibfield{author}{\bibinfo{person}{Samuel Williams}, \bibinfo{person}{Andrew
  Waterman}, {and} \bibinfo{person}{David~A. Patterson}.}
  \bibinfo{year}{2009}\natexlab{}.
\newblock \showarticletitle{Roofline: An Insightful Visual Performance Model
  for Multicore Architectures}.
\newblock \bibinfo{journal}{\emph{Commun. ACM}} \bibinfo{volume}{52},
  \bibinfo{number}{4} (\bibinfo{year}{2009}), \bibinfo{pages}{65--76}.
\newblock


\end{thebibliography}

\appendix
\section{Statistical tests on the scale curve}
\label{app:scale}
\setlength{\intextsep}{2pt plus 2pt minus 1pt}
\setlength{\textfloatsep}{2pt plus 2pt minus 1pt}
\setlength{\floatsep}{2pt plus 2pt minus 1pt}

This appendix carries the tests behind \S\ref{sec:evaluation}. Perplexity
intervals are paired Student $t$ intervals on the mean per-window NLL
difference (12 windows, 11 d.f.), exponentiated; the shared fingerprint
makes window $i$ the same text at every size, so steps and the knee form
window by window on the log ratio. MMLU deltas are subject-stratified
paired bootstraps (10\,000 draws) with exact McNemar over the same 2\,280
questions; between-size steps compose two standard errors in quadrature (no
cross-size pairing). All intervals cover the evaluation corpus, not the
calibration draw, priced by Table~\ref{tab:seeds}.

\begin{table}[htbp]
\centering
\footnotesize
\setlength{\tabcolsep}{4pt}
\begin{tabular}{lrrrrrrl}
\toprule
 & \multicolumn{4}{c}{WikiText-2 perplexity} & \multicolumn{2}{c}{b/param} & MMLU, 4-bit $-$ 2-bit \\
\cmidrule(lr){2-5}\cmidrule(lr){6-7}\cmidrule(lr){8-8}
Model & FP16 & 2-bit $\times$ & excess & fall, \% & 2-bit & 4-bit & gap, points \\
\midrule
Qwen3-4B  & 12.2369 & 1.3845 & 0.3845 & --- & 5.162 & 5.302 & 14.45 \ [11.60, 17.27] \\
Qwen3-8B  & 8.9899  & 1.2201 & 0.2201 & $-42.8$ \ $[-51.8, -33.5]$ & 5.322 & 5.956 & 7.49 \ [5.28, 9.70] \\
Qwen3-14B & 7.9820  & 1.1894 & 0.1894 & $-13.9$ \ $[-22.8, -4.9]$ & 5.106 & 5.404 & 6.09 \ [3.62, 8.52] \\
\bottomrule
\end{tabular}
\caption{Three models, one configuration and harness, nine matching
fingerprints. \emph{Excess} is the perplexity ratio minus one; \emph{fall}
its relative drop from the row above (paired 95\% CI); \emph{b/param} is
whole-model, embedding included; gap cells: paired bootstraps. Data:
echelle-4b-8b.csv, mmlu-appariee.csv, ppl-genou.csv.}
\label{tab:scale}
\end{table}

The nine paired perplexity intervals and MMLU deltas behind
Figure~\ref{fig:scale}'s error bars are in \texttt{ppl-appariee.csv} and
\texttt{mmlu-appariee.csv}; no perplexity interval contains zero, all 108
window-level comparisons agree in sign, and two MMLU intervals (FP16 $-$
4-bit at 4B and 14B) contain zero.

\begin{table}[htbp]
\centering
\footnotesize
\setlength{\tabcolsep}{4pt}
\begin{tabular}{lrrrlc}
\toprule
4B arm & 4B ppl & step 4B$\to$8B ($t$) & step 8B$\to$14B ($t$) & knee [95\% CI] ($t$) & excludes zero \\
\midrule
published & 16.9422 & $-0.1265$ ($-9.62$) & $-0.0255$ ($-3.38$) & $-0.1010$ $[-0.1377, -0.0643]$ ($-6.06$) & yes \\
seed 1    & 16.7425 & $-0.1146$ ($-9.22$) & $-0.0255$ ($-3.38$) & $-0.0891$ $[-0.1274, -0.0509]$ ($-5.13$) & yes \\
seed 2    & 15.8836 & $-0.0619$ ($-5.47$) & $-0.0255$ ($-3.38$) & $-0.0365$ $[-0.0744, +0.0014]$ ($-2.12$) & no \\
seed 3    & 15.1027 & $-0.0115$ ($-0.93$) & $-0.0255$ ($-3.38$) & $+0.0139$ $[-0.0268, +0.0547]$ ($+0.75$) & no \\
\bottomrule
\end{tabular}
\caption{The knee (first step minus second step, per window, in the log
perplexity ratio against FP16) replayed with each 4B calibration re-draw in
place of the published artifact. Conditional on the artifacts as built it
excludes zero; one re-draw moves it from $-0.10$ to $+0.01$, and the third
leaves the first step itself no longer separated from zero. Data:
knee-seeds.csv.}
\label{tab:seeds}
\end{table}

On perplexity the knee of Table~\ref{tab:seeds} excludes zero for the
artifacts as built; on the MMLU gap to 4 bits the steps are $6.96$ points
($\mathrm{SE}$ 1.82, $p = 0.0001$) and $1.40$ ($\mathrm{SE}$ 1.68,
$p = 0.40$), the second supporting neither a flattening nor a constant pace.
Three differences drive the divergence: power (49\,140 paired tokens against
2\,280 unpaired questions), skill dimension (2 bits damages reasoning more
than recall, \S\ref{sec:evaluation}), and reference arm. Against AWQ the
knee is $-0.0576$ $[-0.1011, -0.0140]$, $t = -2.91$, and the 8B$\to$14B step
$-1.58\%$ $[-3.14, -0.004]$, at the interval's edge and not a closing.
Neither supports extrapolation beyond three sizes.

\section{Protocol, pre-registered criteria and provenance}
\label{app:protocol}

\textbf{Layout benchmark protocol.} The ten arms of Table~\ref{tab:layouts}
run in one process on one L40S, interleaved each round in fixed dispatch
order, seven rounds with two discarded. Speedups are medians of per-round
ratios against the FP16 control, with ranges. Two phases (nine arms, then
the same nine plus the 2-bit competitor) move the nine common ratios by at
most 0.36\%. Incumbents reproduce across jobs (Planes14
2.15$\times$ twice, hoisted Golay70 1.77$\times$ and 1.78$\times$). Byte
accounting is the kernel's (payload, exception side-channels, f32 tail
columns, row scales); class tables are excluded as resident constants.
Three uncorrected asymmetries understate our arms: the FP16 control bills
its tail at f16, the competitors none. GB/s is each arm's fastest kept
round; medians leave every percentage unchanged. CUDA events bound the host
share of wall time at 0.1--0.2\%. Pre-registration fixes each criterion
before its measurement; it disciplines the reading and is no substitute for
the cross-job replications above. Runtime: candle 0.9.2, CUDA 12.4.1
images, NVRTC-compiled kernels for the target SM, drivers 580.178.04 (L40S)
and 580.159.03 (A100 clock runs), the 2-bit competitor at its pinned
upstream commit.

\begin{table}[H]
\centering
\scriptsize
\setlength{\tabcolsep}{3pt}
\renewcommand{\arraystretch}{0.85}
\begin{tabular}{lp{0.175\linewidth}p{0.14\linewidth}p{0.135\linewidth}p{0.15\linewidth}p{0.14\linewidth}}
\toprule
 & Ours (4 layouts) & FP16 / cuBLAS & AWQ w4g128 & QTIP 2-bit & No-weights \\
\midrule
Payload & sealed 4B artifact, transcoded bit-exact & same weights, decoded f16 & our weights requantized (timing only) & pseudo-random valid stream (timing only) & none \\
Grid & one warp per row, 252 launches & ours / library's & own, as shipped & own, as shipped (1\,024 threads, 64~KiB smem) & ours \\
Output, accum. & f32, f32 FMA & f32 / binary16 & binary16 & f32 & f32 \\
f64 check & $10^{-5}$ (worst $3.4\times10^{-8}$) & $10^{-5}$ / $10^{-3}$ (worst $5.7\times10^{-5}$) & $10^{-3}$ (worst $5.8\times10^{-5}$) & $10^{-5}$ (worst $5.4\times10^{-8}$) & $10^{-5}$ (exact) \\
Tail, scales billed & f32 tail + f32 row scales & f16 tail in 16.000 & none (all columns quantized) & none (shapes divide) & tail + scales only (0.159 b/w) \\
Registers, local & 40, 0~B & 42, 0~B / n/a & 32, 0~B & 48--56, 0~B & 31, 0~B \\
Quality claim & the artifact's (\S\ref{sec:evaluation}) & reference & none & none & none \\
\bottomrule
\end{tabular}
\caption{What the ten-arm comparison holds equal (matrices, shapes,
process, rounds, f64 verification) and what it does not. Differences are as
shipped and uncorrected; the billing asymmetries understate our arms.}
\label{tab:fairness}
\end{table}

\begin{table}[H]
\centering
\scriptsize
\setlength{\tabcolsep}{3.5pt}
\renewcommand{\arraystretch}{0.8}
\begin{tabular}{p{0.46\linewidth}lp{0.37\linewidth}}
\toprule
Criterion & Fixed before & Outcome \\
\midrule
Golay70: kept only if $\geq 1.6\times$ FP16 &
2026-08-07 &
1.31$\times$ [1.29--1.32]; 1.34$\times$ in the ten-arm run. Discarded. \\
Golay70 hoisted: adopted for the served path only if $\geq 2.0\times$ FP16
and $\geq 20\%$ whole-model memory margin over the deployed 4-bit checkpoint &
2026-08-11 &
Memory margin 23.3\% by arithmetic; speed 1.77$\times$ [1.76--1.78]. Not
adopted. \\
QTIP arm: $r = t(\text{Planes14})/t(\text{QTIP})$ per round; verdict
fixed for a range above 1 (inherited claim confirmed), below 1 (refuted),
covering 1 (undecided) &
2026-08-20 &
2.27$\times$ [2.27--2.28], entirely above 1. \\
QTIP fraction of its byte bound $\leq 59.6\%$ (the no-weights control's
bound, 4.77$\times$) &
2026-08-20 &
61.1\%, 1.5 points above at a fixed 0.2 resolution: the control is a
property of our launch geometry. \\
A100: order of the lattice arms (Planes14 $>$ Planes12x $>$ Slot32 $>$
Golay70) preserved beyond range overlap, else published as a property of the
format--card pair &
2026-08-18 &
Planes14 leads (0.79$\times$), the Golay70 arms trail; Planes12x and
Slot32 tie at 0.73$\times$, so the ordering is published as a property of
the format--card pair (\S\ref{sec:a100}). \\
14B served VRAM reproduces the byte-level 5.106 b/param within $\pm 0.5\%$ &
2026-08-17 &
5.0866 b/param, $-0.38\%$. \\
Fused served path: the two \texttt{LLVQ\_FUSE} arms emit the same 128 tokens
and diverge from the dense engine at the same position; launch counts 144
against 252; memory grows by exactly $4$ bytes a fused row; end-to-end ratio
in [1.00, 1.12] &
2026-08-24 &
All met: same tokens, both diverging at token 89, 144/252,
$+3\,686\,400$ bytes exactly, $1.061\times$ [1.050--1.069]. \\
End-to-end ranges, anomalies A1--A4: divergence at or before token 5;
tokens differing between rounds of one arm; median outside $\pm 10\%$ of
the earlier point; range above 10\% of the median &
2026-08-18 &
None triggered: 4B divergence at token 89 (the tie-break of the earlier run),
worst shift $-1.6\%$, worst range 0.6\%. \\
\bottomrule
\end{tabular}
\caption{Decision criteria and predictions fixed before the measurement
that tested them. Dates are those of the committed pre-registration files,
several with OpenTimestamps proofs; the first row predates that practice.
Outcomes are read from the journals of Table~\ref{tab:provenance}.}
\label{tab:prereg}
\end{table}

\begin{table}[H]
\centering
\scriptsize
\setlength{\tabcolsep}{2.5pt}
\renewcommand{\arraystretch}{0.75}
\renewcommand{\UrlBreaks}{\do\-\do\/\do\_}%
\renewcommand{\UrlBigBreaks}{}%
\begin{tabular}{>{\raggedright\arraybackslash}p{0.115\linewidth}>{\raggedright\arraybackslash}p{0.355\linewidth}>{\raggedright\arraybackslash}p{0.47\linewidth}}
\toprule
Figure / Table & CSV (\path{docs/data/}) & Journal (\path{docs/mesures/}) \\
\midrule
Fig.~\ref{fig:records}, Fig.~\ref{fig:layouts}, Table~\ref{tab:layouts} & \path{echelle-formats.csv}; field widths from the kernel headers (\path{llvq-cuda/kernels/*.cuh}) & \path{f2-p3-qtip-banc-2026-08-21.txt} \\
Fig.~\ref{fig:dataflow} & --- (kernel source, \path{llvq-cuda/kernels/planes.cu}, \path{llvq_planes.cuh}, \path{qtip_prelude.cuh}) & \path{f2-p3-qtip-banc-2026-08-21.txt} (registers, spill) \\
Fig.~\ref{fig:attribution} & \path{attribution-slot32.csv} & \path{attribution-cuda-2026-08-05.txt}, \path{fusion-qkv-cuda-2026-08-05.txt}, \path{f2-p3-qtip-banc-2026-08-21.txt} (l.~260--295) \\
Fig.~\ref{fig:a100}, \S\ref{sec:a100} clocks & \path{echelle-formats.csv}, \path{echelle-formats-a100.csv} & \path{f4-a100-2026-08-18.txt}, \path{g-horloges-planes12x-2026-08-23.txt} \\
Planes12x served (\S\ref{sec:evaluation}) & --- & \path{g-horloges-planes12x-2026-08-23.txt} \\
Fused served path (\S\ref{sec:attribution}) & --- & \path{d1-fusion-servie-2026-08-24.txt} \\
Fig.~\ref{fig:scale} & \path{ppl-appariee.csv}, \path{mmlu-appariee.csv} (interval cells), \path{echelle-4b-8b.csv} & \path{ppl-appariee-4b-2026-08-17.txt}, \path{ppl-appariee-8b-14b-2026-08-17.txt}, \path{mmlupair-4b-8b-2026-08-13.txt}, \path{mmlupair-14b-2026-08-17.txt} \\
Table~\ref{tab:phases} & \path{phases.csv} & \path{phases-2026-08-07.txt} \\
Table~\ref{tab:e2e} & \path{campagne-finale.csv}, \path{tableau-8b.csv}; the 14B row from the journal & \path{b2-fusedrun-plages-2026-08-18.txt} \\
Tables~\ref{tab:campaign}, \ref{tab:campaign8b} & \path{campagne-finale.csv}, \path{tableau-8b.csv} & \path{a4-campagne-2026-08-06.txt}, \path{campagne-finale-bras4-2026-08-07.txt}, \path{campagne-8b-qualite-2026-08-08.txt}, \path{campagne-8b-q8-2026-08-08.txt}, \path{b2-fusedrun-plages-2026-08-18.txt} \\
Table~\ref{tab:lit} & --- & transcription of the original paper's tables (\path{docs/llvq-paper-notes.md}) \\
Table~\ref{tab:envelope} & --- & \path{fusedrun-14b-2026-08-17.txt} \\
Table~\ref{tab:scale} & \path{echelle-4b-8b.csv}, \path{mmlu-appariee.csv}, \path{ppl-genou.csv} & \path{rtbits-planes-8b-2026-08-09.txt}, \path{rtbits-14b-2026-08-17.txt}, \path{campagne-14b-qualite-2026-08-10.txt} \\
Table~\ref{tab:seeds} & \path{knee-seeds.csv} & \path{f5-graines-4b-2026-08-19.txt}, \path{a4-campagne-4b-ppl-BRUT-2026-08-06.txt} \\
Table~\ref{tab:prereg} & --- & \path{proofs/preregistration-*.md} \\
\bottomrule
\end{tabular}
\caption{Where each figure and table comes from. Every CSV is committed;
the build regenerates the figures and fails the checked tables on drift.
Token fingerprints matched across quality arms (\S\ref{sec:evaluation}):
\texttt{3f1baca9033bf251} (perplexity), \texttt{65dcd53655e8bfa5} (MMLU).
Rented GPU time cost under \$100; per-job amounts are in \texttt{jobs.csv}.}
\label{tab:provenance}
\end{table}

\end{document}